%% file: main_neurips_submission.tex
\documentclass{article}

\usepackage[main, final]{neurips_2026}

\usepackage{times}

\usepackage{multicol}
\usepackage{graphicx}
\usepackage{float}
\usepackage{amsmath}
\usepackage{amssymb}
\usepackage{array}
\usepackage{booktabs}
\usepackage{multicol,multirow}
\usepackage{tabularx}
\usepackage{setspace}
\usepackage{makecell}
\usepackage{fontawesome}
\usepackage[dvipsnames]{xcolor}
\usepackage{tasks}
\usepackage{pifont}
\usepackage{colortbl}
\usepackage{xcolor}
\usepackage{framed}
\usepackage{wrapfig}  
\usepackage{titletoc}
\usepackage{color} 
\usepackage{caption, subcaption, overpic, textpos}
\usepackage{titlesec}
\usepackage{comment}

\makeatletter
\DeclareRobustCommand\onedot{\futurelet\@let@token\@onedot}
\def\@onedot{\ifx\@let@token.\else.\null\fi\xspace}

\makeatother

\usepackage{xspace}
\newcommand{\modelname}{SparseVideoNav\xspace}

\definecolor{deemph}{gray}{0.6}

\definecolor{baselinecolor}{gray}{.9}
\definecolor{yellow}{RGB}{218,165,32}
\definecolor{lightcyan}{rgb}{0.88, 1.0, 1.0}
\definecolor{lightskyblue}{rgb}{0.53, 0.81, 0.98}
\definecolor{aliceblue}{rgb}{0.94, 0.97, 1.0}
\definecolor{LightSlateBlue}{RGB}{70,130,180}
\definecolor{DeepBlue}{RGB}{65,100,170}
\definecolor{DeepPurple}{RGB}{136,105,160}
\definecolor{LightGreen}{RGB}{59,125,35}
\definecolor{LightRed}{RGB}{234,66,53}
\definecolor{cvprblue}{rgb}{0.21,0.49,0.74}

\usepackage[pagebackref,breaklinks,colorlinks,citecolor=cvprblue,bookmarks=true]{hyperref}

\usepackage[capitalize]{cleveref}

\crefname{section}{Sec.}{Secs.}
\crefname{section}{Sec.}{Secs.}
\crefname{figure}{Fig.}{Figs.} % For lowercase references (e.g., "fig. 1")
\Crefname{figure}{Fig.}{Figs.} % For uppercase references (e.g., "Fig. 1")

\crefname{table}{Tab.}{Tabs.} % For lowercase references (e.g., "tab. 1")
\Crefname{table}{Tab.}{Tabs.} % For uppercase references (e.g., "Tab. 1")

\usepackage{epigraph}
\newcommand{\underfigtab}{\vspace{-10pt}}

\newlength\savewidth

\renewcommand{\paragraph}[1]{\vspace{1.25mm}\noindent\textbf{#1}}

\newcolumntype{x}[1]{>{\centering\arraybackslash}p{#1pt}}
\newcolumntype{y}[1]{>{\raggedright\arraybackslash}p{#1pt}}
\newcolumntype{z}[1]{>{\raggedleft\arraybackslash}p{#1pt}}

\newcommand{\app}{\raise.17ex\hbox{$\scriptstyle\sim$}}

\usepackage[utf8]{inputenc} % allow utf-8 input
\usepackage[T1]{fontenc}    % use 8-bit T1 fonts
\usepackage{hyperref}       % hyperlinks
\usepackage{url}            % simple URL typesetting
\usepackage{booktabs}       % professional-quality tables
\usepackage{amsfonts}       % blackboard math symbols
\usepackage{nicefrac}       % compact symbols for 1/2, etc.
\usepackage{microtype}      % microtypography
\usepackage{xcolor}         % colors

\title{Sparse Video Generation
Propels Real-World Beyond-the-View Vision-Language Navigation}

\author{%
  Hai Zhang\thanks{Equal contribution} \\
  \And
  Siqi Liang\footnotemark[1] \\
  \And
  Li Chen \\
  \And
  Yuxian Li \\
  \AND % <--- 强制换行，把后面4个人分到下一行
  Yukuan Xu \\
  \And
  Yichao Zhong \\
  \And
  Fu Zhang \\
  \And
  Hongyang Li \\
  \And
  \\
  The University of Hong Kong \\
  \texttt{\url{https://github.com/OpenDriveLab/SparseVideoNav}} 
}

\begin{document}

\noindent

\maketitle

\vspace{-5mm}
\begin{center}
    \centering
    \captionsetup{type=figure}
    \includegraphics[width=0.95\textwidth]{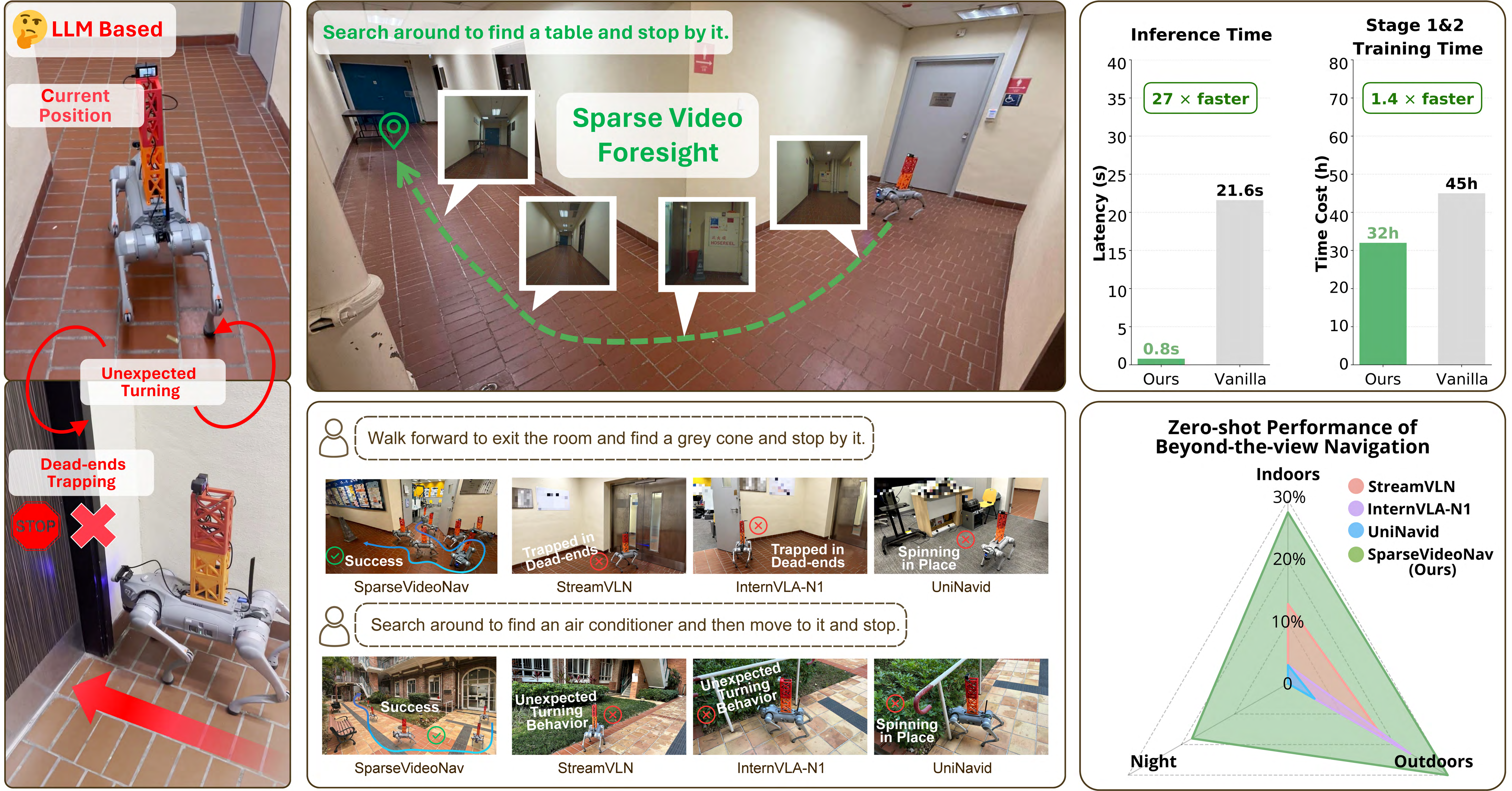}
    % \vspace{-6pt}
    \captionof{figure}{In this work, we investigate beyond-the-view navigation task in the real world, where agents must locate distant, unseen targets without step-by-step guidance.
    Traditional large language model-based methods suffer from short-horizon supervision, leading to short-sighted behaviors, \textit{e.g.}, unexpected turning and dead-end trapping. 
    To address this challenge, we pioneer the use of video generation model as the primary backbone for action generation.
    % We then sparsify 
    The whole training pipeline
    is sparsified further for the sake of extended prediction horizon and computational efficiency.
    % Compared to the unoptimized vanilla model, we achieves a $27\times$ speed-up in inference latency and a $1.4\times$ speed-up in training efficiency.
    % Extensive real-world zero-shot experiments demonstrate that we achieves $2.5\times$ the success rate of state-of-the-art LLM baselines on beyond-the-view tasks and marks the first realization of such capability in challenging night scenes.
    \label{fig:teaser}
    \underfigtab
    }
    \vspace{5mm}
\end{center}%

\begin{abstract}
  Why must vision-language navigation be bound to detailed and verbose language instructions?
While such details ease decision-making, they fundamentally contradict the goal for navigation in the real-world.
Ideally, agents should possess the autonomy to navigate in unknown environments guided solely by simple and high-level intents. 
Realizing this ambition introduces a formidable challenge: Beyond-the-View Navigation (BVN), where agents must locate unseen targets without step-by-step guidance.
Existing large language model (LLM)-based methods, though adept at following dense instructions, often suffer from short-sighted behaviors due to their reliance on short-horizon supervision.
Simply extending the supervision horizon, however, destabilizes LLM training.
In this work, we identify that video generation models (VGM) inherently benefit from long-horizon supervision to align with language instructions, rendering them uniquely suitable for BVN tasks.
Capitalizing on this insight, we propose applying VGM as the main backbone to generate actions for the first time in this field.
Yet, the prohibitive latency for generating videos spanning tens of seconds makes real-world deployment impractical.
To bridge this gap, we propose \modelname, achieving sub-second trajectory inference guided by a generated sparse future spanning a 20-second horizon.
This yields a remarkable 27$\times$ speed-up compared to the unoptimized counterpart.
Extensive real-world zero-shot experiments demonstrate that \modelname achieves 2.5$\times$ the success rate of state-of-the-art LLM baselines on BVN tasks and marks the first realization of such capability in challenging night scenes.
\end{abstract}

\input{Introduction/introduction}

\input{RelatedWork/relatedwork}

\input{Method/method}
\input{Evaluations/evaluations}

\input{Conclusion/conclusion}
\section*{Acknowledgment}
This study is supported by National Natural Science Foundation of China (62206172). 
This work is in part supported by the JC STEM Lab of Autonomous Intelligent Systems funded by The Hong Kong Jockey Club Charities Trust. We are grateful to Jiazhi Yang and Di Zhang for their valuable discussions, 
% and we thank 
Haoguang Mai, Longyang Wu, Hongchen Li, Shuo Diao, Haolin Ou, Ronghao Li, and members from OpenDriveLab for their assistance throughout this project. We also 
acknowledges DJI for funding support during the project.

% % \newpage
{
\bibliographystyle{plainnat}
\bibliography{bibliography_short, bibliography_custom}
}

\appendix
%\input{paper-template-latex/rebuttal_response}
% \maketitlesupplementary
\title{
\huge{Sparse Video Generation
Propels Real-World Beyond-the-View Vision-Language Navigation} \\
\textit{Supplementary Material}
}

% % You will get a Paper-ID when submitting a pdf file to the conference system
% \author{Author Names Omitted for Anonymous Review. Paper-ID 
% [16]}

\input{Appendix/appendix_submission}

\newpage
\input{checklist.tex}

\end{document}

%% file: Introduction/introduction.tex
\section{Introduction}
\label{sec:intro}
\vspace{-3pt}

Vision-language navigation (VLN) empowers agents to perform complex tasks by grounding natural language instructions into sequential actions based on visual observations~\citep{ku-etal-2020-room, zhang2026compass1}. 
Recently, the advent of large-language models (LLMs) has catalyzed significant breakthroughs in this field~\citep{zhang2024uni, zhang2025embodied, wei2025ground}. 
However, a fundamental tension remains: while real-world interactions typically demand navigation based on simple and high-level intents, current agents often require dense and step-by-step instructions~\citep{cheng2024navila, wei2025streamvln}, a paradigm typically termed Instruction-Following Navigation (IFN). 
This clear discrepancy highlights a formidable but less-explored frontier: \textit{Beyond-the-View Navigation} (BVN), where agents must autonomously locate unseen targets in the absence of such granular and intermediate guidance.

The primary bottleneck in BVN for existing LLM-based methods is their inherent short sight. 
These models are typically supervised by short-horizon actions (\textit{e.g.}, 4 to 8 steps) during training~\citep{wei2025streamvln, zhang2024uni, internvla-n1}. 
Consequently, they struggle to infer correct navigational intents, typically causing two failure modes during deployment.
First, the inability to observe the target over a long distance induces severe uncertainty, resulting in unexpected turning behaviors or spinning in place.
Second, upon entering a dead end, the agent mistakenly assumes the path end, leading to dead-end trapping.
While extending the supervision horizon seems like a logical fix, it often destabilizes the training process of LLM~\citep{wang2025trackvla}, rendering it a non-viable solution.

In this work, we pivot toward video generation models (VGMs) based on a key observation: unlike LLMs, VGMs are inherently pre-trained to capture long-horizon future aligned with the language instruction~\citep{wan2025, wu2025hunyuanvideo}. 
Recognizing this strength, we identify VGMs as a well-suited interface to provide the long-horizon foresight that BVN demands. 
Nevertheless, whether we should strictly adhere to the standard VGM paradigm of predicting \textit{\textbf{continuous}} videos~\citep{hu2025video, internvla-n1} remains a question worthy of scrutiny.
We contend that the dense and high-frequency temporal information required by continuity is redundant for guiding navigation. 
This insight drives us to explore a \textit{\textbf{sparse}} video generation paradigm. 
By doing so, we aim to achieve the dual advantage of extending the prediction horizon while simultaneously reducing both training and inference overhead.
To this end, we reformulate the training objective by employing sparse video frames corresponding to strategically selected timesteps as the direct supervision signal for VGM.

Translating this sparse foresight into a comprehensive navigation system, yet, presents two challenges.
Primarily, the computational overhead of injecting the whole history, coupled with the extensive denoising steps required to generate future predictions, still leads to significant inference latency that makes real-world deployment impractical.
In addition, unlike LLM-based methods that can bridge the sim-to-real gap by co-training with heterogeneous real-world data~\citep{zhang2024uni, wei2025streamvln}, VGM lacks a straightforward mechanism to leverage such data sources. 

To bridge this gap, we introduce \modelname, a \textbf{Sparse Video} generation-based system for vision-language \textbf{Nav}igation.
For the efficiency bottleneck, we instantiate \modelname via a structured training pipeline.
For the data challenge, we curate a large-scale navigation dataset spanning 140 hours across diverse real-world environments.
Through zero-shot evaluations in six real-world scenes, we demonstrate that \modelname achieves state-of-the-art (SOTA) performance on BVN tasks, marking the first realization of such capability in challenging night scenes.

\textbf{Contributions:} 

\textbf{1)} We investigate beyond-the-view navigation tasks in the real world from a new perspective by applying video generation model as the primary backbone for action generation. 
Crucially, we empirically identify that supervising with longer horizon is instrumental in improving performance.

\textbf{2)} We advance beyond the conventional constraint of continuous video generation in existing methods by pioneering a paradigm shift towards sparse video generation. This reformulation liberates the capability to reason over substantially longer prediction horizon while achieving $1.4\times$ training speed-up and $1.7\times$ inference speed-up.

\textbf{3)} \modelname achieves sub-second trajectory inference guided by a generated sparse future spanning a 20-second horizon. 
This yields a remarkable $27\times$ speed-up compared to the unoptimized counterpart.
Extensive real-world zero-shot experiments demonstrate \modelname adeptly handles beyond-the-view tasks even in complex terrains like dead ends, narrow accessible ramp, and hillside with high inclination angles, achieving $2.5\times$ the success rate of SOTA LLM baselines
% in beyond-the-view scenarios, 
and marking the first realization of such capability in challenging night scenes with a $17.5\%$ success rate.

\textbf{4)} We open-source a real-world navigation dataset spanning over 140 hours, representing the largest-scale real-world navigation dataset to date.

\vspace{-3pt}

%% file: RelatedWork/relatedwork.tex
\section{Related Work}
\vspace{-3pt}

\subsection{Foundation Models for Vision-Language Navigation}
\vspace{-3pt}
The integration of foundation models into VLN predominantly transitions from modular, training-free approaches to end-to-end fine-tuning approaches.
Training-free approaches leverage the zero-shot reasoning capabilities of off-the-shelf LLMs to perform specific sub-tasks, such as task reasoning~\citep{long2024instructnav}, stage scheduling~\citep{cao2025cognav}, frontier detection~\citep{gong2025stairway}, and object validation~\citep{yu2024vlngame}.
While these modular approaches maintain interpretability, they inherently suffer from cascading error propagation~\citep{yu2024vlngame} or limited generalization capability~\citep{yokoyama2024vlfm, chen2025position}.

To better optimize spatio-temporal information throughout the training pipeline, fine-tuning approaches have gained increasing traction, which aim to map multimodal inputs directly to actions in an end-to-end manner~\citep{zhang2024uni, zhang2025novel}.
Recent SOTA methods have demonstrated strong generalization through large-scale training.
Notably,~\citet{zhang2024uni} proposes Uni-Navid to establish a competitive baseline by unifying diverse navigation tasks with a single versatile policy.
StreamVLN~\citep{wei2025streamvln} introduces a streaming framework that manages continuous video inputs via hybrid slow-fast context modeling to accelerate the inference. 
Despite their promise, these methods predominantly require dense and step-by-step instructions, a dependency that contradicts with practical real-world demands. 
When facing beyond-the-view tasks guided by simple and high-level intents, they are prone to falter due to the vulnerability induced by short-horizon supervision. 
Consequently, there is a pressing need for a novel paradigm designed to bridge this gap.

\subsection{Video Generation Models for Embodied Agents}
\vspace{-3pt}

Distinguished from text-centric pretraining, VGM implicitly encodes dynamic variation between frames, leading to a smaller domain gap towards downstream actions.
This unique property has driven an explosion of interest across a wide spectrum of embodied tasks from robotic manipulation~\citep{bu2024clover, hu2025video, ye2026worldactionmodelszeroshot} to autonomous driving~\citep{gao2024magicdrivedit, yang2025resim, zhao2025drivedreamer}.
Within the specialized scope of navigation, InternVLA-N1~\citep{internvla-n1} demonstrates that co-training with video generation objectives can improve downstream policy performance.
However, its primary backbone remains LLM, making it susceptible to the limitations of short-horizon supervision.
To the best of our knowledge, \modelname is the first to apply VGM as the primary backbone for action generation in VLN. 
Unlike Navigation World Model~\citep{bar2025navigation}, which is restricted to predicting the immediate next frame based on low-level action inputs, our framework is purely language-conditioned, directly processing high-level intents to achieve long-horizon video foresight.
Concurrent work~\citep{huang2026navdreamervideomodelszeroshot} also uses VGM as a zero-shot high-level navigator, but does not train the VGM backbone to further improve the performance.

Noticeably, these endeavors adopt the default prerequsite to generate continuous videos.
While some concurrent works in manipulation explore sparsification to balance action and video tokens~\citep{motus2026, lingbotva2026}, their future predictions are typically limited to a mere few seconds. 
In contrast, we are the first to harness the sparsification for significantly extending the prediction span (\textit{e.g.}, 20 seconds). 
This enables efficient long-horizon supervision, ultimately unlocking substantial performance improvements in BVN tasks.

\vspace{-3pt}

\begin{figure*}[t]
    \centering
    \includegraphics[width=0.9\linewidth]{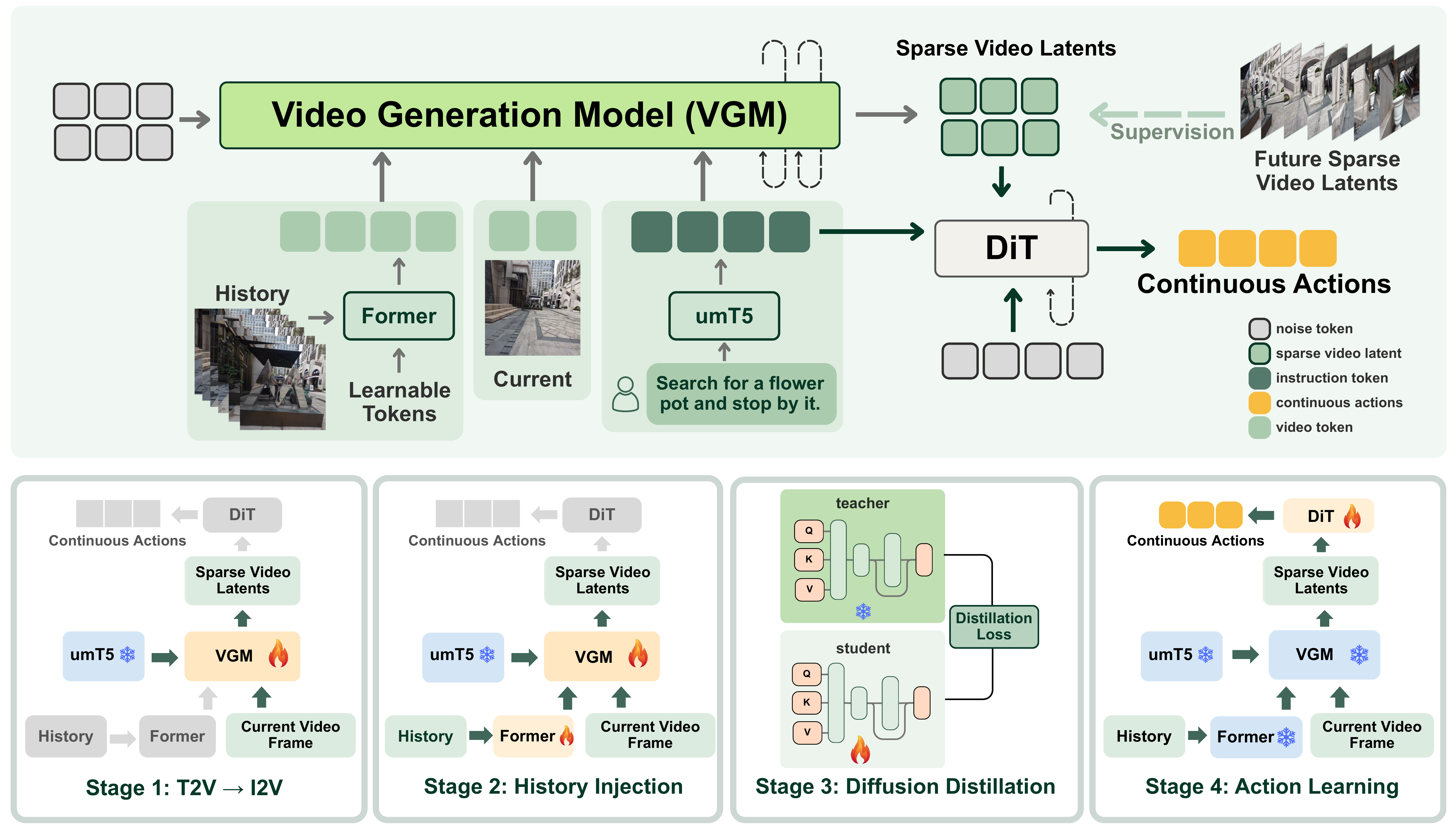}
    \caption{
    \textbf{Architecture and four-stage training pipeline of \modelname}. \textbf{(Top)} denotes our whole training architecture. Current observation, historical observations, and the language instruction are fed into the video generation model (VGM) backbone to generate future sparse video latents.
    DiT-based action head then predicts continuous actions conditioned on generated sparse future and the language instruction. \textbf{(Bottom)} denotes our four-stage training pipeline, 
    with \textbf{Stage 1 (\Cref{sec:stage1})} adapting T2V to I2V, 
    \textbf{Stage 2 (\Cref{sec:stage2})} injecting history into I2V backbone; 
    \textbf{Stage 3 (\Cref{sec:stage3})} distilling the backbone to reduce denoising steps; 
    \textbf{Stage 4 (\Cref{sec:stage4})} learning actions based on generated sparse future.
    Components not utilized in a specific stage are indicated by \textcolor{gray}{gray blocks}.
    }
    \vspace{-7pt}
    \label{fig:model_architecture}
\end{figure*}

%% file: Method/method.tex
\section{Methodology}
\label{sec:Methodology}
\begin{wrapfigure}[19]{r}{0.44\textwidth}
    \vspace{-4em}
    \centering
    \includegraphics[width=0.44\textwidth]{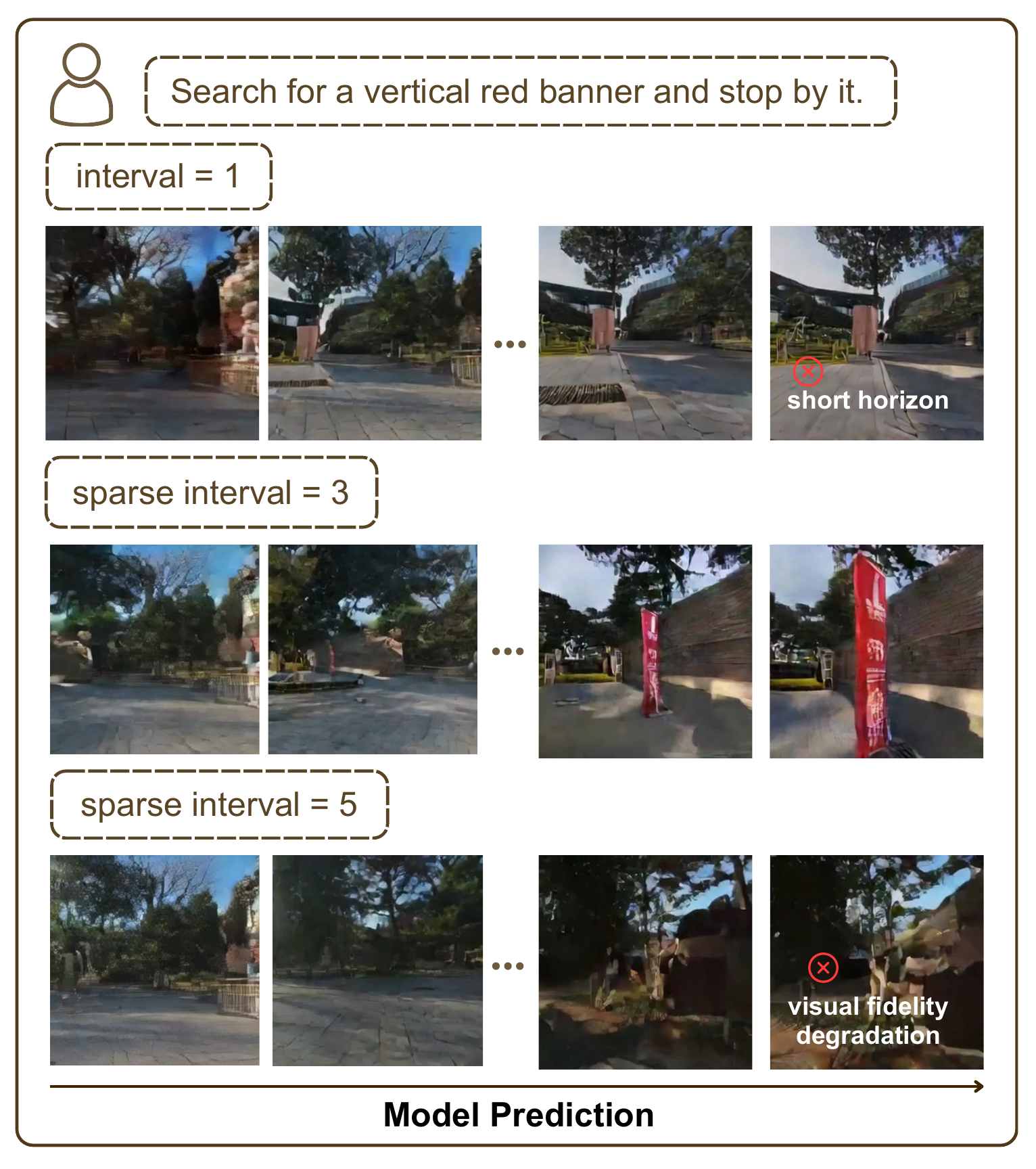}
    \caption{\textbf{Qualitative comparison of different sparse intervals.} With the sparse interval of 3, the model successfully imagines a path towards beyond-the-view target, while maintaining visual fidelity.}
    \label{fig:sparse_design}
    % \vspace{-2em}
\end{wrapfigure}
\vspace{-3pt}

The design of \modelname is guided by two fundamental pillars: a training paradigm-level innovation through sparsification 
{(\Cref{sec:sparsification})}, and a system-level innovation via 
synergistic orchestration.
To achieve this, we develop a data curation pipeline 
{(\Cref{sec:data_curation})} alongside a structured four-stage 
training pipeline.
The linchpin of the training pipeline is the \textit{sparse future video}, which serves as the supervision across stages 1, 2, and 3, and the conditional input of stage 4.

The whole architecture and training pipeline are shown in \Cref{fig:model_architecture}, where \texttt{Former} block denotes Q-Former~\citep{li2023blip} and Video-Former~\citep{hu2025video} used to compress history.

\subsection{Sparsification}
\vspace{-3pt}
\label{sec:sparsification}
To extend the prediction horizon, we bring in sparse video supervision into the training of VGM to enable fixed-interval video generation.
As shown in \Cref{fig:sparse_design}, setting the interval to 3 strikes the best balance between the prediction horizon and the visual fidelity.
To further ensure action prediction accuracy, we maintain continuous generation for the first two observation chunks, covering 8 timesteps.
This design leads to the sparse generation timesteps at $[T+1, T+2, T+5, T+8, T+11, T+14, T+17, T+20]$, spanning 20 seconds at 4 FPS, where $T$ denotes the current timestep.

\subsection{Data Curation Pipeline}
\label{sec:data_curation}
\vspace{-3pt}

Different from methods built upon video-based LLMs~\citep{zhang2024uni, zhang2025embodied}, which can mitigate the sim-to-real gap by co-training on pure simulation navigation data~\citep{savva2019habitat} mixed with real-world VQA datasets, VGM cannot effectively leverage this strategy. 
Relying exclusively on simulation data typically leads to mode collapse~\citep{srivastava2017veegan, thanh2020catastrophic}. 
Furthermore, existing real-world navigation datasets are often characterized by severe fisheye distortion~\citep{shah2021rapid, hirose2023sacson} and limited scale~\citep{hirose2018gonet}, making them unsuitable for fine-tuning VGM.

To address these challenges, we develop a data curation pipeline involving human operators equipped with a handheld camera to capture diverse videos.
% , as shown in~\Cref{fig:data_curation}. 
To minimize human jitter that blocks VGM to learn consistent dynamics, we explicitly utilize DJI Osmo Action 4 with RockSteady+ stabilization.

We collect 140 hours of real-world navigation videos. 
These videos are processed into approximately 13,000 trajectories via uniform temporal sampling, with an average length of 140 frames at 4 FPS. 
Subsequently, we estimate camera poses using Depth Anything 3 (DA3)~\citep{depthanything3} to extract continuous action labels. 
Detailed visualizations concerning action extraction are deferred in~\Cref{sec:data_curation_details}.
Finally, language instructions are manually annotated by human experts. 
This pipeline establishes the largest real-world VLN dataset to date.
We will release this dataset to benefit the community.

% \begin{figure}[t!]
%     \centering
%     \includegraphics[width=1.\linewidth]{paper-template-latex/figures/data_curation.pdf}
%     \caption{\textbf{Data curation pipeline.} Our data curation pipeline transforms raw human-collected videos into paired ego-centric frames, annotated instructions, and continuous action labels.
%     \HY{
%     Do we exhibit novelty here? Or simply state the facts, if yes, put it in the Appendix.}
%     }
%     \label{fig:data_curation}
%     \vspace{-10pt}
% \end{figure}

\subsection{Stage 1: T2V $\rightarrow$ I2V }
\label{sec:stage1}
\vspace{-3pt}

To achieve a balance between computational overhead and video generation fidelity, we adopt Wan2.1-1.3B T2V~\citep{wan2025} model as our backbone. 
Wan utilizes a 3D causal VAE~\citep{kingmaauto} structure to compress the spatio-temporal dimensions.
Specifically, given an input video $V\in \mathbb{R}^{(1+T)\times H\times W\times 3}$, Wan-VAE encodes it into latent chunks $C\in \mathbb{R}^{1+T/4, H/8, W/8, 16}$, resulting in $(1+T/4)$ chunks where the shape of each chunk is $[H/8, W/8, 16]$.
By default, all subsequent training processes are performed at the chunk level.

Note that the T2V model is designed to generate the future mainly based on language instructions rather than visual inputs. 
Hence, the first stage involves adapting the backbone from T2V to image-to-video (I2V) to ensure the consistency of the generated future with the initial observation.

We retain the original flow matching~\citep{lipman2023flow} objective of Wan for fine-tuning.
During training, given a chunk latent at an arbitrary timestep $c_T$, the following sparse chunk latents $x_1=[c_{T+1}, c_{T+2}, c_{T+5}, c_{T+8}, ..., c_{T+20}]$, a random noise $x_0\sim \mathcal{N}(0,I)$, and a timestep $t\in [0,1]$ sampled from a logit-normal distribution, an intermediate latent $x_t$ is obtained as the training input.
$x_t$ is defined as a linear interpolation between $x_0$ and $x_1$:
\begin{align}
    x_t = tx_1 + (1-t)x_0.
\end{align}
The ground-truth velocity $v_t$ is defined as:
\begin{align}
    v_t = \frac{dx_t}{dt} = x_1 - x_0.
\end{align}
The loss function is formulated as mean square error between the model output and the velocity $v_t$:
\begin{align}
    \mathcal{L}_{\text{stage1}} = \mathbb{E}_{x_0, x_1, l, c_T, t}||u(x_t, l, c_T, t; \theta) - v_t ||^2,
\end{align}
where $l$ is the language embedding of umT5~\citep{chung2023unimax}, $\theta$ is the model weights, and $u(x_t, l, c_T, t; \theta)$ denotes the predicted velocity.

\subsection{Stage 2: History Injection}
\label{sec:stage2}
\vspace{-3pt}

A critical distinction between navigation foundation models and previous vision-language-action models lies in the necessity of incorporating the entire history of observations~\citep{kim2024openvla, bu2025univla, jiang2025wholebodyvla, wang2026nativememnativememorycompression}. 
Unlike LLMs that can directly absorb a long sequence of image tokens, VGMs lack this capability to process this input~\citep{wan2025, wu2025hunyuanvideo}.
We draw inspiration from CDiT~\citep{bar2025navigation} architecture to inject the history information with efficient training and inference computational overhead. 
Specifically, we introduce an additional cross-attention block within each transformer block of Wan backbone to explicitly inject history information.
To preserve the generative priors of the fine-tuned I2V model, we initialize the final linear layer of these newly added cross-attention blocks with zeros.

Nevertheless, the whole history input presents another challenge due to its excessive length and high dimensionality.
To efficiently extract spatio-temporal features, we employ a two-step strategy that utilizes a Q-Former~\citep{li2023blip} to process features along the temporal dimension and subsequently applies a Video-Former~\citep{hu2025video} for features along the spatial dimension.

We denote the history embedding processed by Q-Former and Video-Former at an arbitrary timestep as $h_T$, the training objective is formulated as:
\begin{align}
    \mathcal{L}_{\text{stage2}} = \mathbb{E}_{x_0, x_1, l, c_T, h_T, t}||u(x_t, l, c_T, h_T, t; \theta) - v_t ||^2.
\end{align}

\subsection{Stage 3: Diffusion Distillation}
\label{sec:stage3}
\vspace{-3pt}

Unlike manipulation tasks, which involve limited visual changes~\citep{wu2024unleashing, kim2024openvla, shi2025diversity} and allow high-fidelity reconstruction with few denoising steps~\citep{hu2025video, bu2025agibot_iros}, navigation tasks involve highly dynamic scene transitions~\citep{zhang2024uni, zhang2025embodied}. 
Consequently, generating high-fidelity future in navigation scenarios via few-step denoising is inherently difficult, posing significant challenges for real-world deployment. 
While video generation-based methods have been explored in autonomous driving~\citep{gao2024magicdrivedit, zhao2025drivedreamer}, they still suffer from high inference latency, requiring tens of seconds to minutes for generation. 

To address this efficiency bottleneck, we adapt PCM~\citep{wang2024phased} to flow-matching paradigm for distilling history-injected I2V model.
In this setup, we designate the history-injected I2V model as the teacher model and initialize an architecturally identical student model with the same weights.
We then partition the noise schedule into 4 phases, where the student model learns to predict the solution point of each phase on the probability flow ODE~\citep{song2021scorebased, song2023consistency} trajectory of the teacher model~\citep{wang2024phased}. 
By minimizing the consistency loss between adjacent timesteps, we progressively distill the inference steps from $N=50$ to $M=4$.

\subsection{Stage 4: Action Learning}
\label{sec:stage4}
\vspace{-3pt}

Drawing inspiration from VPP~\citep{hu2025video}, we freeze the distilled I2V model and employ an inverse dynamics paradigm to predict continuous actions. Specifically, the generated sparse future and the language instruction are injected into the DiT-based action head via cross-attention.
Nevertheless, we observe a clear visual discrepancy between the generated and the original ground-truth future frames, which leads to a misalignment between the synthesized dynamics and the original action labels. 
To combat this inconsistency, we employ DA3 to relabel the generated future frames, thereby ensuring the action supervision is precisely aligned.

% Though we utilize PCM to reduce the computational overhead during the denoising process, performing on-the-fly video generation during action training remains infeasible. 
% To mitigate this issue, we preprocess and then cache the generated sparse future observations for all timesteps. 

We adopt DDIM~\citep{song2021denoising} to reconstruct the relabeled actions $\overline{a}_0$ from noised action $\overline{a}_k=\sqrt{\overline{\beta}_k \overline{a}_0} + \sqrt{1 - \overline{\beta}_k}\epsilon $, where $\epsilon$ denotes white noise and $\overline{\beta}_k$ is the noisy coefficient at timestep $k$.
The training process can be formulated as learning a denoised $D_\psi$:
\begin{align}
    \mathcal{L}_{action}(\psi; A) = \mathbb{E}_{\overline{a}_0,\epsilon,k}||D_\psi(\overline{a}_k, l, \overline{V}) - \overline{a}_0||^2,
\end{align}
where $\overline{V}$ denotes the generated sparse latent future observation.

\vspace{-3pt}

%% file: Evaluations/evaluations.tex
\section{Evaluations}
\label{sec:exp}
\vspace{-3pt}

To validate the efficacy of \modelname, we design our experiments to investigate: \textbf{1)} Can \modelname, as a framework grounded in video generation, exhibit foundational competency in following language instructions? (See \Cref{sec:main_resuls}); \textbf{2)} Does the sparse foresight successfully translate to superior performance in challenging beyond-the-view scenarios? (See \Cref{sec:main_resuls} and \Cref{sec:ablation}); \textbf{3)} Does our architectural design strikes a good balance between efficiency and performance? (See \Cref{sec:ablation}); \textbf{4)} Can \modelname demonstrate clear scalability with increased data, adaptability to dynamic obstacles, and robustness to camera height variations? (See \Cref{sec:ablation} and \Cref{sec:further_discussion})

% \begin{figure}[htbp]
%     \centering
%     \begin{minipage}[c]{0.3\textwidth}
%         \centering
%         \includegraphics[width=\linewidth]{figures/bad_scene.pdf} 
%         \caption{\textbf{Scenes of Habitat.}}
%         \label{fig:habitat_bad_scenes}
%     \end{minipage}\hfill
%     \begin{minipage}[c]{0.66\textwidth}
%         \centering
%         \captionof{table}{\textbf{Quantitative results of all methods on R2R-CE subset.}}
%         \label{tab:vlnce}
%         \resizebox{\linewidth}{!}{%
%         \begin{tabular}{lcccc}
%         \toprule
%         \multirow{2}{*}{Method} & \multicolumn{4}{c}{R2R Val-Unseen} \\
%         \cmidrule(lr){2-5}
%          & NE$\downarrow$ & OS$\uparrow$ & SR$\uparrow$ & SPL$\uparrow$ \\
%         \midrule
%         Uni-NaVid~\citep{zhang2024uni} & 5.29 & 55.36 & 45.31 & 41.19 \\
%         InternVLA-N1~\citep{internvla-n1} & 5.53 & 59.37 & 48.95 & 42.59 \\
%         StreamVLN~\citep{wei2025streamvln} & 4.97 & 60.64 & 50.01 & 41.50 \\
%         \midrule
%         \rowcolor{blue!10}
%         SparseVideoNav & \textbf{3.80} & \textbf{62.62} & \textbf{53.34} &  \textbf{50.21} \\
%         \bottomrule
%         \end{tabular}
%         }
%     \end{minipage}
% \end{figure}

\subsection{Experimental Setup}
\label{sec:experiment_setup}
\vspace{-3pt}

\noindent\textbf{Baselines.} To demonstrate the superiority of the VGM backbone, we choose three SOTA LLM-based models for comparison: \textbf{1)}Uni-Navid~\citep{zhang2024uni} uses video-based LLM to unify diverse VLN tasks; \textbf{2)}StreamVLN~\citep{wei2025streamvln} is designed to accelerate the inference through KV cache. \textbf{3)}InternVLA-N1~\citep{internvla-n1} uses the video generation objective to co-train the backbone.

\begin{table*}[t]
\centering
\small
\caption{\textbf{Quantitative results of zero-shot performance on diverse real-world scenes.} \modelname achieves unanimously SOTA performance across all the scenes on both instruction-following navigation (IFN) and beyond-the-view navigation (BVN) tasks. The highest success rate across \modelname and all baselines is highlighted in \textbf{bold}. (Numbers represent \textbf{success rate} in \%)}
\label{tab:main_results}
\resizebox{0.95\textwidth}{!}{
\setlength{\tabcolsep}{5pt} 
\renewcommand{\arraystretch}{1.} 

\begin{tabular}{@{} l cc cc cc cc cc cc | cc @{}}
\toprule
& \multicolumn{4}{c}{Indoors} & \multicolumn{4}{c}{Outdoors} & \multicolumn{4}{c|}{Night} & \multicolumn{2}{c}{\textbf{Average}} \\
\cmidrule(lr){2-5} \cmidrule(lr){6-9} \cmidrule(lr){10-13} \cmidrule(l){14-15}
& \multicolumn{2}{c}{Room} & \multicolumn{2}{c}{Lab Bldg} & \multicolumn{2}{c}{Yard} & \multicolumn{2}{c}{Park} & \multicolumn{2}{c}{Square} & \multicolumn{2}{c|}{Mountain} & \multicolumn{2}{c}{Total} \\
\cmidrule(lr){2-3} \cmidrule(lr){4-5} \cmidrule(lr){6-7} \cmidrule(lr){8-9} \cmidrule(lr){10-11} \cmidrule(lr){12-13} \cmidrule(l){14-15}
\textbf{Method} & IFN & BVN & IFN & BVN & IFN & BVN & IFN & BVN & IFN & BVN & IFN & BVN & IFN & BVN \\
\midrule
\rowcolor{gray!5} \textit{Baselines} &&&&&&&&&&&&&& \\
Uni-NaVid~\citep{zhang2024uni} & 20 & 5 & 10 & 0 & 10 & 0 & 20 & 10 & 0 & 0 & 0 & 0 & 10.0 & 2.5 \\
StreamVLN~\citep{wei2025streamvln} & 45 & 10 & 40 & 15 & 30 & 5 & 50 & 30 & 20 & 0 & \textbf{25} & 0 & 35.0 & 10.0 \\
InternVLA-N1~\citep{internvla-n1} & 25 & 5 & 15 & 0 & 35 & 5 & 30 & \textbf{40} & 0 & 0 & 0 & 0 & 17.5 & 8.3 \\
\midrule
\rowcolor{blue!5} \textbf{SparseVideoNav} & \textbf{55} & \textbf{25} & \textbf{55} & \textbf{30} & \textbf{50} & \textbf{20} & \textbf{65} & \textbf{40} & \textbf{50} & \textbf{20} & \textbf{25} & \textbf{15} & \textbf{50.0} & \textbf{25.0} \\
\midrule
\rowcolor{gray!5} \textit{Ablation Study} &&&&&&&&&&&&&& \\
(a) \textit{w distilled 4 steps cont. 2} & 25 & 0 & 20 & 0 & 10 & 0 & 35 & 15 & 5 & 0 & 0 & 0 & 15.8 & 2.5 \\
(b) \textit{w distilled 4 steps cont. 10} & 50 & 10 & 40 & 5 & 35 & 5 & 45 & 40 & 30 & 5 & 20 & 5 & 36.7 & 11.7 \\
(c) \textit{w/o distilled 50 steps cont. 20} & 70 & 35 & 65 & 45 & 55 & 30 & 75 & 60 & 75 & 20 & 35 & 25 & 62.5 & 35.8 \\
(d) \textit{\modelname w/o Former} & 40 & 15 & 60 & 25 & 40 & 20 & 60 & 45 & 45 & 20 & 25 & 10 & 45.0 & 22.5 \\
\bottomrule
\end{tabular}
}

\vspace{0pt} % 增加上下两块内容的间距

% 中间的同行图表 (左图，右表)
\begin{minipage}[c]{0.3\textwidth}
    \centering
    \includegraphics[width=\linewidth]{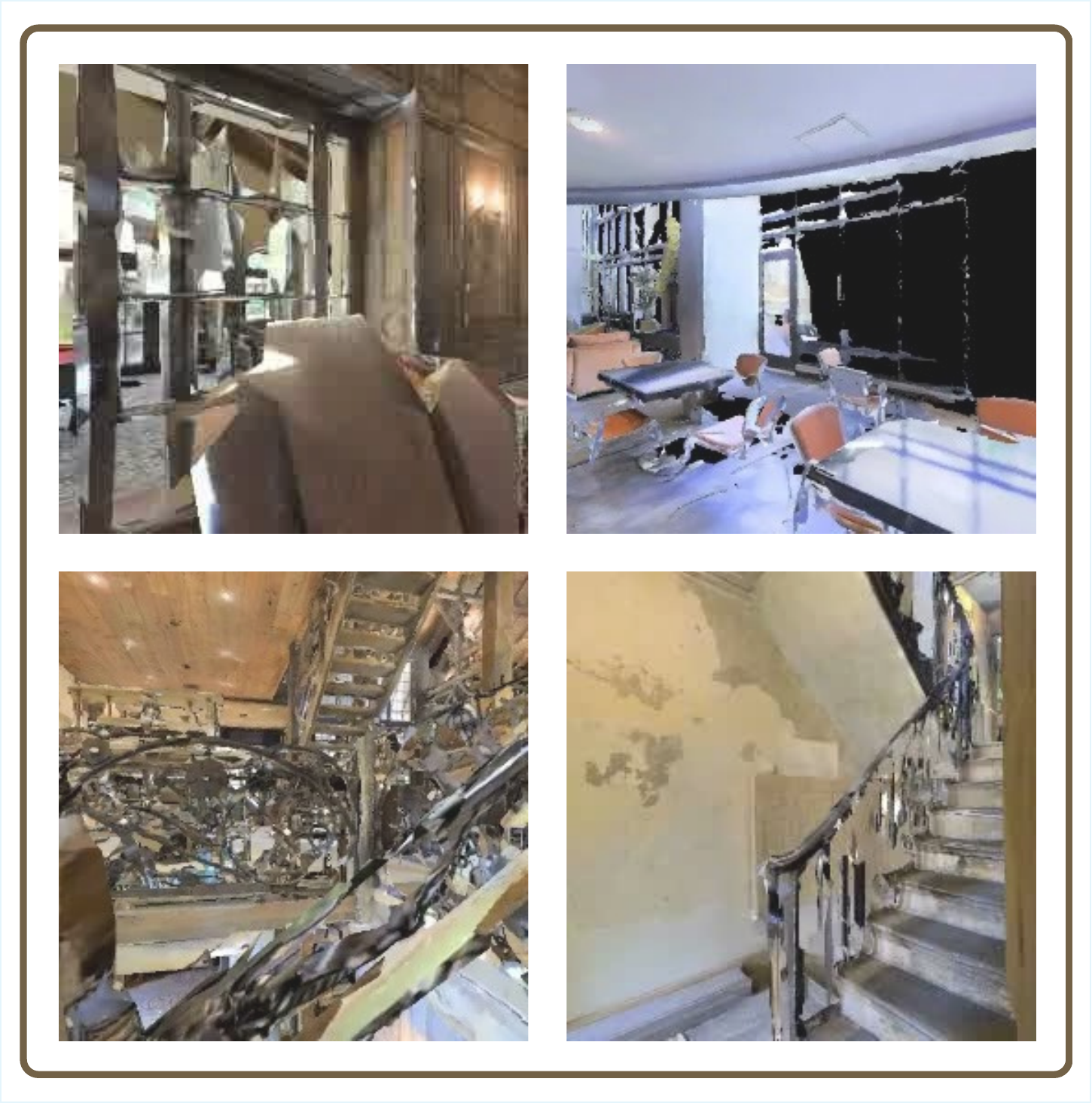} 
    \captionof{figure}{\textbf{Scenes of Habitat.}}
    \label{fig:habitat_bad_scenes}
\end{minipage}\hfill
\begin{minipage}[c]{0.66\textwidth}
    \centering
    \captionof{table}{\textbf{Quantitative results of all methods on R2R-CE.}}
    \label{tab:vlnce}
    \resizebox{0.9\linewidth}{!}{%
    \begin{tabular}{lcccc}
    \toprule
    \multirow{2}{*}{Method} & \multicolumn{4}{c}{R2R Val-Unseen} \\
    \cmidrule(lr){2-5}
     & NE$\downarrow$ & OS$\uparrow$ & SR$\uparrow$ & SPL$\uparrow$ \\
    \midrule
    Uni-NaVid~\citep{zhang2024uni} & 5.58 & 53.3 & 47.0 & 42.7 \\
    InternVLA-N1~\citep{internvla-n1} & 4.72 & 60.4 & 52.0 & 45.6 \\
    StreamVLN~\citep{wei2025streamvln} & 4.89 & 60.6 & 54.4 & 52.1  \\
    \midrule
    \rowcolor{blue!10}
    SparseVideoNav & 4.17 & 56.1 & 48.6 &  47.4 \\
    \bottomrule
    \end{tabular}
    }
\end{minipage}
\vspace{-10pt}
\end{table*}

% ================= 浮动体 2：单独的底部大图 =================
\begin{figure*}[t]
    \centering
    \includegraphics[width=0.9\linewidth]{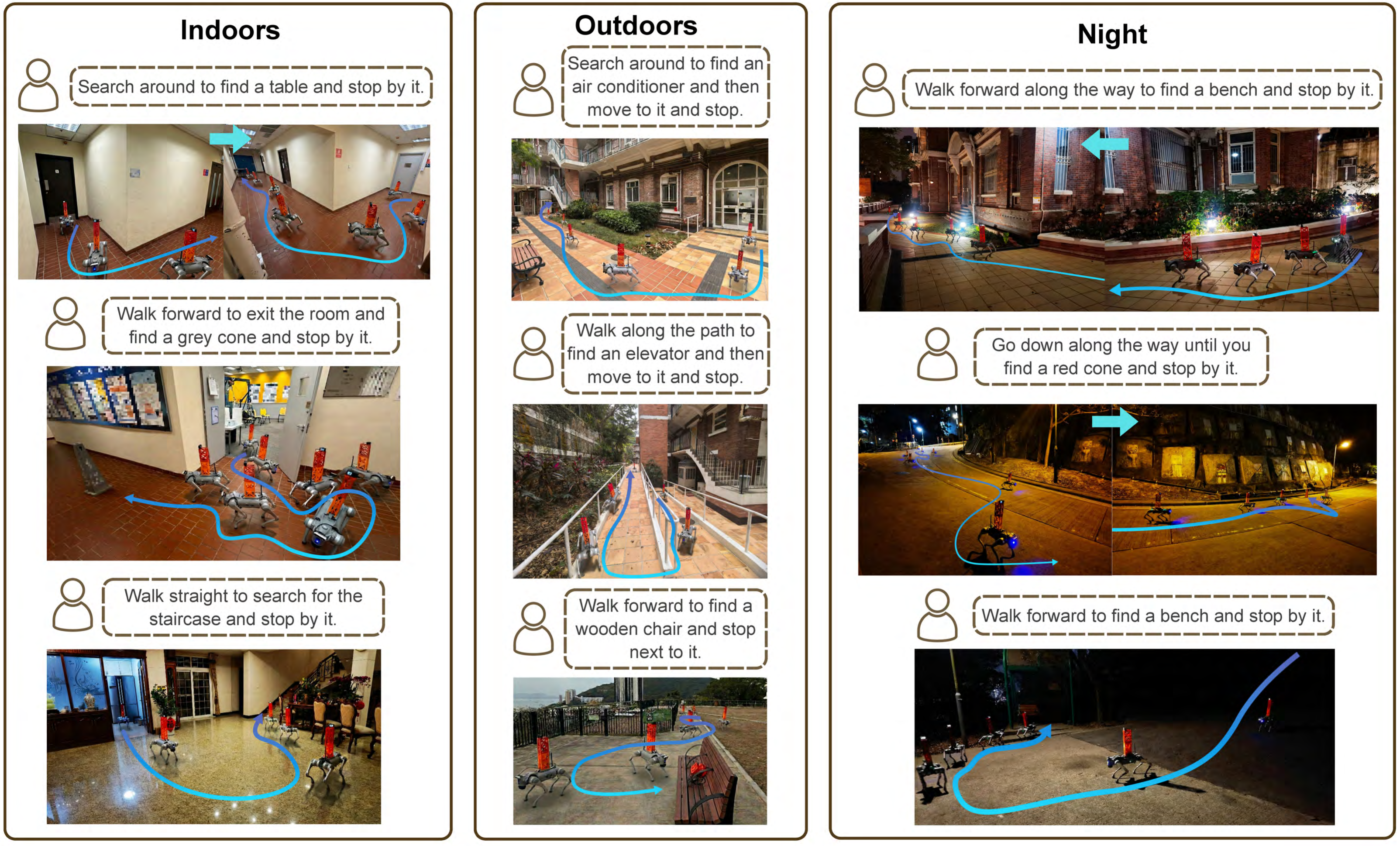}
    \caption{\textbf{Qualitative results of zero-shot beyond-the-view navigation in challenging, unstructured environments.} 
    \modelname successfully navigates through challenging scenarios, including dead ends, narrow accessible ramp, and hillside with high inclination angles.
    }
    \label{fig:beyond_the_view_nav_vis}
    \vspace{-10pt}
\end{figure*}

\noindent\textbf{Evaluation Protocols.}
To assess the navigation capabilities in real-world environments, we select six diverse unseen scenes across three categories: \textbf{Indoors} (Room, Lab Building), \textbf{Outdoors} (Yard, Park), and \textbf{Night} (Square, Mountain) to test the \textbf{zero-shot} generalization capability. 
For each scene, we design four distinct navigation tasks, consisting of two standard instruction-following navigation (IFN) tasks and two challenging beyond-the-view navigation (BVN) tasks. 
Detailed specifications for all the tasks are deferred in 
\Cref{sec:task_specifications}.
To ensure statistical reliability, each model is tested 10 times per task, resulting in a total of \textbf{240 trials} per model. 
We report the \textbf{Success Rate} averaged across task types for each scene.
We observe that LLM-based baselines often stop with a lateral orientation towards the target, whereas our model typically stops facing the target directly.
To ensure a fair comparison, we define success solely based on proximity, where a trial is considered successful if the agent stops within 1.5 meters of the target.
To minimize the influence of time and weather, we conduct comparative tests for all models within the same time window to ensure consistent environmental conditions.
Hardware deployment details are deferred in~\Cref{sec:hardware_deployment_details}.

To assess foundational language understanding capabilities in the established benchmark, we leverage the VLN-CE R2R~\citep{r2r2022}. 
We only use R2R, RxR~\citep{ku2020room}, R2R-EnvDrop-CE~\citep{tan2019learning}, and ScaleVLN~\citep{cruz2023scaling} to train our model from scratch without any dagger data.

% \begin{figure*}[t!]
%     \centering
%     \includegraphics[width=0.8\linewidth]{figures/vgm_pred.pdf}
%     \caption{\textbf{Analysis of video generation results of \modelname} during zero-shot deployment in beyond-the-view navigation.}
%     \label{fig:vgm_pred}
%     \vspace{-10pt} 
% \end{figure*}

\begin{figure*}[t] 
    \centering
    
    \begin{minipage}[t]{0.6\linewidth}
        \centering   
        \vspace{0pt} 
        \includegraphics[width=\linewidth]{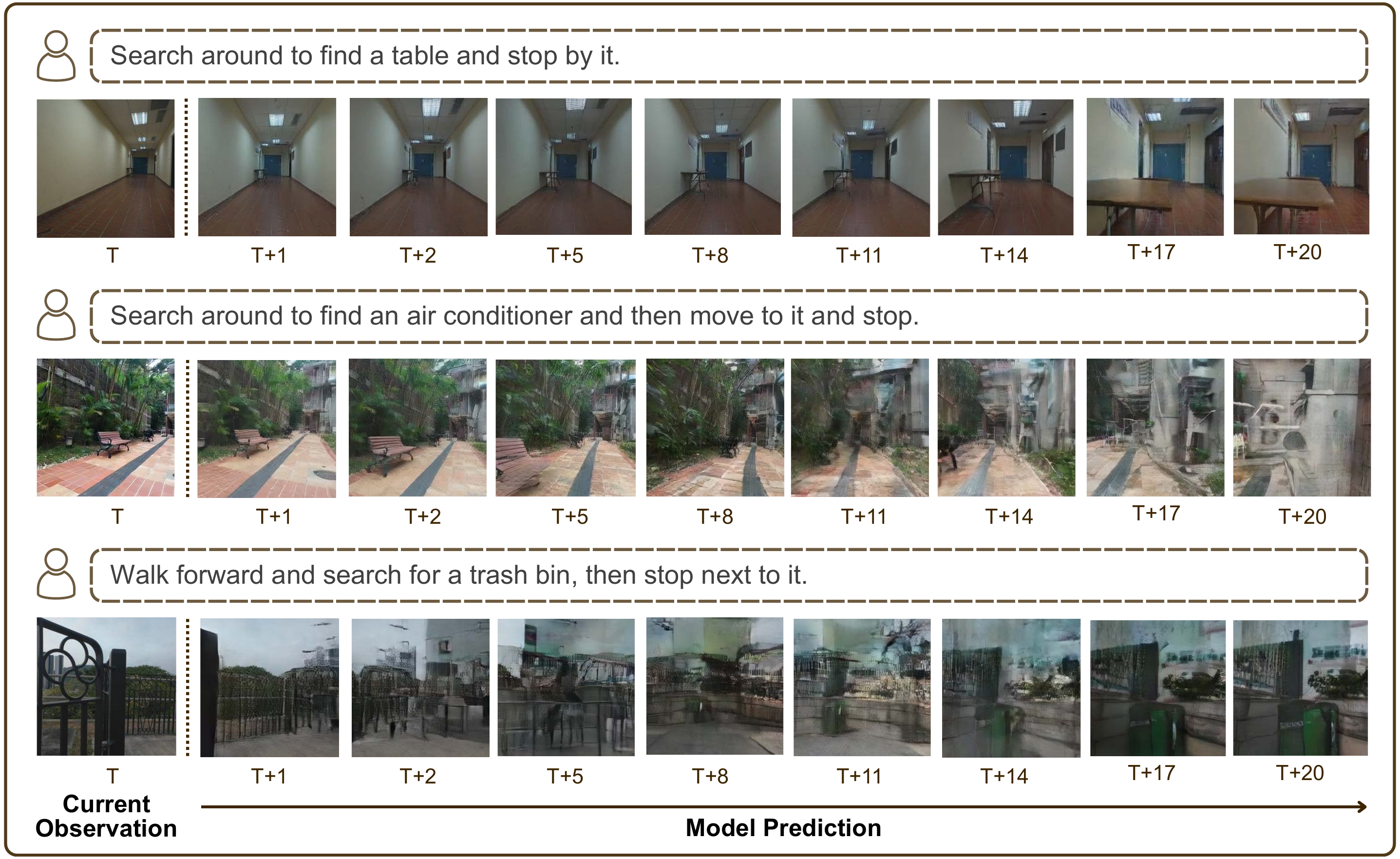}
        \caption{\textbf{Analysis of video generation results} during zero-shot deployment in beyond-the-view navigation.}
        \label{fig:vgm_pred}
    \end{minipage}
    \hfill 
    \begin{minipage}[t]{0.37\linewidth}
        \centering  
        \vspace{0pt} 
        \includegraphics[width=\linewidth]{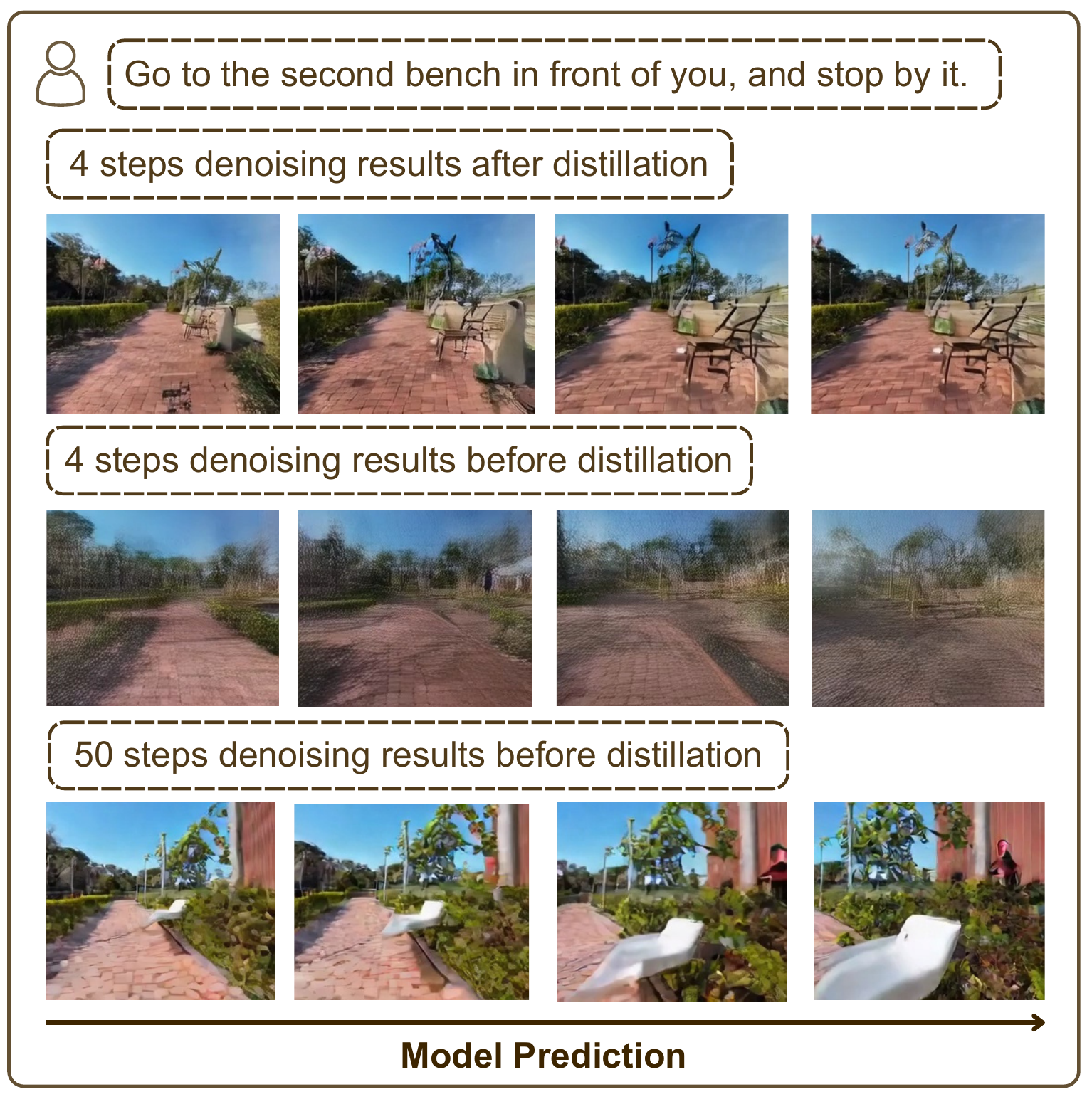}
        \caption{\textbf{Ablation study on diffusion distillation (Stage 3).} 
        % Our distillation strategy enables the model to achieve visual fidelity with only 4 denoising steps (top row) comparable to the original model using 50 steps (bottom row).
        }
        \label{fig:exp_ablation_distillation}
    \end{minipage}
    
    \vspace{-15pt}
\end{figure*}

% \vspace{-10pt}

\subsection{Main Results}
\label{sec:main_resuls}
\vspace{-3pt}

As shown in Table~\ref{tab:main_results}, \modelname achieves unanimously SOTA zero-shot performance across all real-world scenes on both IFN and BVN tasks.
In high-fidelity real-world scenes, the true potential of VGM can be fully unleashed.
Compared to the strongest baseline StreamVLN, \modelname achieves a significant performance improvement, with the average success rate by \textbf{+15.0\%} on IFN and \textbf{+15.0\%} on challenging BVN. 
% This substantial gap emphasizes the effectiveness of our proposed video generation paradigm in handling real-world navigation demands.
Notably, the diminished visibility in challenging night environments exacerbates the short-sighted limitation, precipitating a systematic failure across all established baselines on BVN tasks. 
In contrast, by leveraging robust long-horizon guidance, \modelname stands as the sole method capable of navigating to these distant goals in such extreme environments.
Furthermore, as shown in~\Cref{sec:controlled_real_world}, even when baselines are fine-tuned on our self-collected dataset, they still show no significant performance improvement, demonstrating that the superiority of \modelname stems from its well-designed architecture.

% \Cref{tab:main_results}
As shown in Table~\ref{tab:vlnce}, \modelname achieves reasonable performance on VLN-CE R2R even without any dagger data.
We observe that the poor rendering quality of Habitat~\citep{savva2019habitat} introduces a severe visual gap, as shown in~\Cref{fig:habitat_bad_scenes}. 
Unlike LLM-based methods that map observations into abstract \textbf{semantic} spaces, VGMs are intrinsically more sensitive to visual fidelity as they rely on \textbf{pixel-level reconstruction}. 
Since Wan is pre-trained on high-fidelity videos~\citep{wan2025}, forcing it to reconstruct low-quality scenes would constrain its pre-trained priors. 
Nevertheless, \modelname exhibits the smallest discrepancy between success rate and success weighted by path length among all baselines. 
This minimal gap underscores the profound advantage brought by the long-horizon foresight when guiding the action generation.

\noindent\textbf{Qualitative Results.}
To further showcase the superiority of \modelname in BVN, we visualize several trajectories in~\Cref{fig:beyond_the_view_nav_vis}.
\modelname successfully navigates through these challenging scenarios, including dead ends, narrow accessible ramp, and hillside with high inclination angles.

\noindent\textbf{Analysis.}
To better understand why \modelname excels in BVN tasks, we present model prediction results (See \Cref{fig:vgm_pred}) during deployment and several representative failure cases of prior methods (See ~\Cref{fig:teaser}).
As LLM-based baselines rely on short-horizon supervision, they suffer from inherent short-sighted limitations, leading to unexpected turning under long-range uncertainty and premature trapping in dead ends.
Conversely, by integrating this guidance with closed-loop feedback, \modelname can effectively mitigate this limitation.
To further inspire future explorations, we detail failure cases of \modelname in~\Cref{sec:motivating_questions}.

% \vspace{-10pt}

\subsection{Ablation Study}
\label{sec:ablation}
\vspace{-3pt}

\begin{wrapfigure}[19]{r}{0.5\textwidth}
    \centering
    \includegraphics[width=0.5\textwidth]{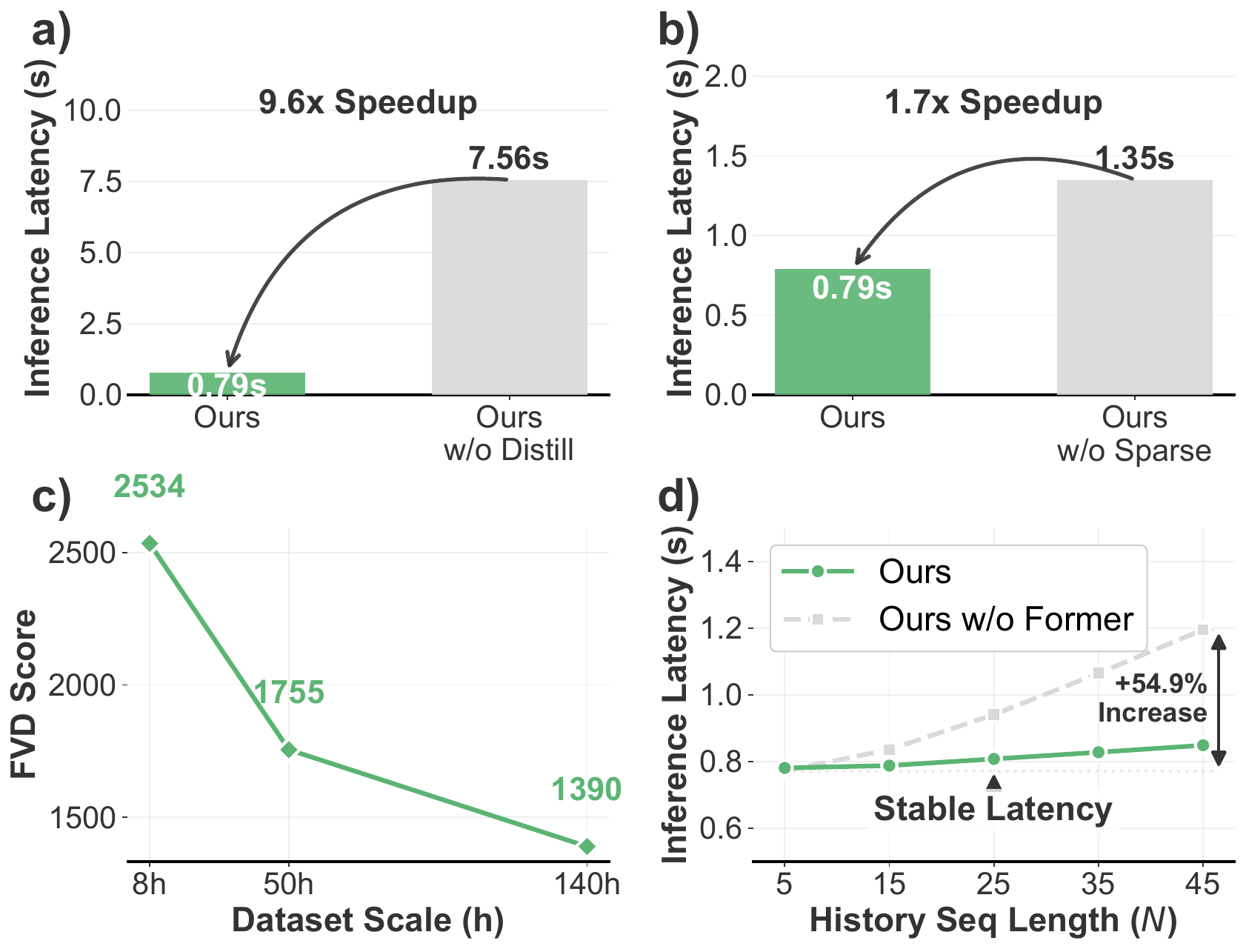}
    \caption{\textbf{Ablation study}
        on a) data scalability and computational overhead comparison over b) sparse design, c) distillation, and d) history compression.}
    \label{fig:ablation_latency_fvd}
    \vspace{-10pt}
\end{wrapfigure}

\noindent\textbf{Diffusion Distillation.}
As evidenced by \Cref{fig:exp_ablation_distillation}, distillation enables VGM to match the visual fidelity of the original 50-step inference with only 4 denoising steps. 
Crucially, as illustrated in \Cref{fig:ablation_latency_fvd}(a), the inclusion of distillation yields a substantial computational advantage, 
accelerating inference by approximately $10\times$ with slight performance compromise (See \textit{Ablation Study} of Table~\ref{tab:main_results} variant (c)). 

\noindent\textbf{Sparse Video Generation.}
To demonstrate the advantage of our proposed sparse design, we devise the following variants for comparative analysis: 

\textbf{a)} \textit{w distilled 4 steps cont. 2} denotes distilled 4-step variant for generating 2 continuous chunks. Variant (a) is specifically designed to emulate the effects of short-sighted limitations typically observed in LLM baselines.

\textbf{b)} \textit{w distilled 4 steps cont. 10} denotes distilled 4-step variant for generating 10 continuous chunks.

\textbf{c)} \textit{w/o distilled 50 steps cont. 20} denotes undistilled 50-step variant for generating 20 continuous chunks, where variant (c) can be seen as an oracle compared to \modelname.

\textbf{d)} \textit{\modelname w/o Former} denotes our method without Q-Former and Video-Former.

As shown in the \textit{Ablation Study} of Table~\ref{tab:main_results}, constrained by the same short-horizon supervision issue as LLM baselines, variant (a) yields suboptimal performance. 
While extending the horizon in variant (b) offers partial mitigation, a distinct performance discrepancy remains compared to \modelname. 
In further comparison with variant (c), \modelname demonstrates substantial efficiency gains, specifically a $1.7\times$ increase in inference speed (See~\Cref{fig:ablation_latency_fvd}(b)) and a 1.4$\times$ reduction in the cumulative convergence time across training stages 1 and 2 (See ~\Cref{fig:teaser}). 
These results show a favorable efficiency-performance trade-off with minimal compromise.

% \begin{figure*}[t] 
%     \centering
%     \begin{minipage}[t]{0.48\linewidth}
%         \vspace{0pt} 
%         \centering
%         \includegraphics[width=0.97\linewidth]{figures/exp_ablation_distillation.pdf}
%         \caption{\textbf{Ablation study on the effectiveness of diffusion distillation (Stage 3).} Our distillation strategy enables the model to achieve visual fidelity with only 4 denoising steps (top row) comparable to the original model using 50 steps (bottom row).}
%         \label{fig:exp_ablation_distillation}
%     \end{minipage}
%     \hfill 
%     \begin{minipage}[t]{0.48\linewidth}
%         \vspace{0pt}
%         \centering
        
%         \includegraphics[width=0.97\linewidth]{figures/people_avoidance.pdf}
%         \caption{\textbf{Discussion of \modelname under dynamic pedestrian interference.} \modelname successfully avoids the oncoming pedestrian and reaches the door.}
%         \label{fig:dynamic_avoidance}
        
%         \vspace{3pt} 
        
%         \includegraphics[width=0.96\linewidth]{figures/low_height.pdf}
%         \caption{\textbf{Discussion of \modelname with the camera fixed at 50cm.} Images in green boxes demonstrate the predictions generated by \modelname. \modelname exhibits strong robustness against height variations.}
%         \label{fig:low_height}
        
%     \end{minipage}
%     \vspace{-10pt}
% \end{figure*}

\noindent\textbf{Data Scalability.}
We conduct the whole training process except Stage 4 on varying data scales: 8h, 50h, and 140h, with additional 3h of unseen data as a validation set to compute FVD~\citep{unterthiner2019fvd}.
As shown in \Cref{fig:ablation_latency_fvd}{(c)}, the consistent downward trend of the FVD curve highlights the scalability of \modelname, demonstrating the capability to absorb large-scale real-world navigation data.

\noindent\textbf{History Compression Strategy.}~\Cref{fig:ablation_latency_fvd}(d) presents the inference latency across varying history lengths.
Former structure decouples latency from history length, ensuring stable inference latency. 
In contrast to the variant without Former, which incurs a $+54.9\%$ latency penalty at $N=45$, \modelname achieves superior efficiency and slightly better performance (See \textit{Ablation Study} of Table~\ref{tab:main_results} variant (d)).

\noindent\textbf{Pretraining from T2V to I2V.} 
To validate the necessity of Stage 1, we benchmark the training time required for convergence under two variants: \textbf{a)} our proposed progressive adaptation; \textbf{b)} training Stage 2 directly from scratch. Experiments are conducted on a cluster of 32 NVIDIA H200 GPUs. The results show that variant (b) requires 64 hours to reach convergence, whereas our progressive adaptation strategy converges in 32 hours. This yields a $2\times$ training speedup with similar performance, demonstrating that Stage 1 serves as an efficient initialization to significantly reduce the overall computational overhead.

\vspace{-10pt}

%% file: Conclusion/conclusion.tex
\section{Conclusion and Limitation} 
\label{sec:conclusion}
\vspace{-10pt}

We introduce \modelname, pioneering the use of VGM as the primary backbone for action generation in VLN.
By shifting to long-horizon sparse foresight, it overcomes short-sightedness and reduces computational overhead, significantly outperforming SOTA LLM baselines in zero-shot real-world evaluations. 
Despite its promise, limitations exist: \textbf{1)} Our 140-hour dataset is not web-scale~\citep{bar2025navigation, mei2025urbannav}, requiring future scaling. \textbf{2)} Inference latency slightly trails LLM paradigms~\citep{wei2025streamvln}, necessitating accelerated distillation and quantization for VGM. 

%% file: Appendix/appendix_submission.tex
\renewcommand{\thetable}{A-\Roman{table}}
\renewcommand{\thefigure}{A-\arabic{figure}}
\setcounter{figure}{0}
\setcounter{table}{0}  
\setcounter{section}{0}

\clearpage
\newpage
\onecolumn
\appendix

\section{Appendix}

In the Appendix, we first compile ``motivating'' questions
% potentially raised from reading the main text 
in~\Cref{sec:motivating_questions}.
% , and we provide detailed explanations for each query. 
Task specifications details in~\Cref{sec:task_specifications} are presented to supplement~\Cref{sec:experiment_setup} in the main paper.
We then provide additional elaborations on data curation details, implementation details and hardware deployment details in~\Cref{sec:data_curation_details},~\Cref{sec:implementation_details} and~\Cref{sec:hardware_deployment_details} respectively. 
Next, we discuss the adaptability, robustness of \modelname and other experiments in~\Cref{sec:further_discussion}.
The Appendix concludes with~\Cref{sec:data_ethics}, which details the ethics statement and licenses.

\subsection{Motivating Questions}
\label{sec:motivating_questions}

\textbf{Q1}: \textit{What are the detailed failure cases of \modelname?}
\vspace{0.3\baselineskip}

\begin{figure*}[h!]
    \centering
w    \includegraphics[width=0.7\linewidth]{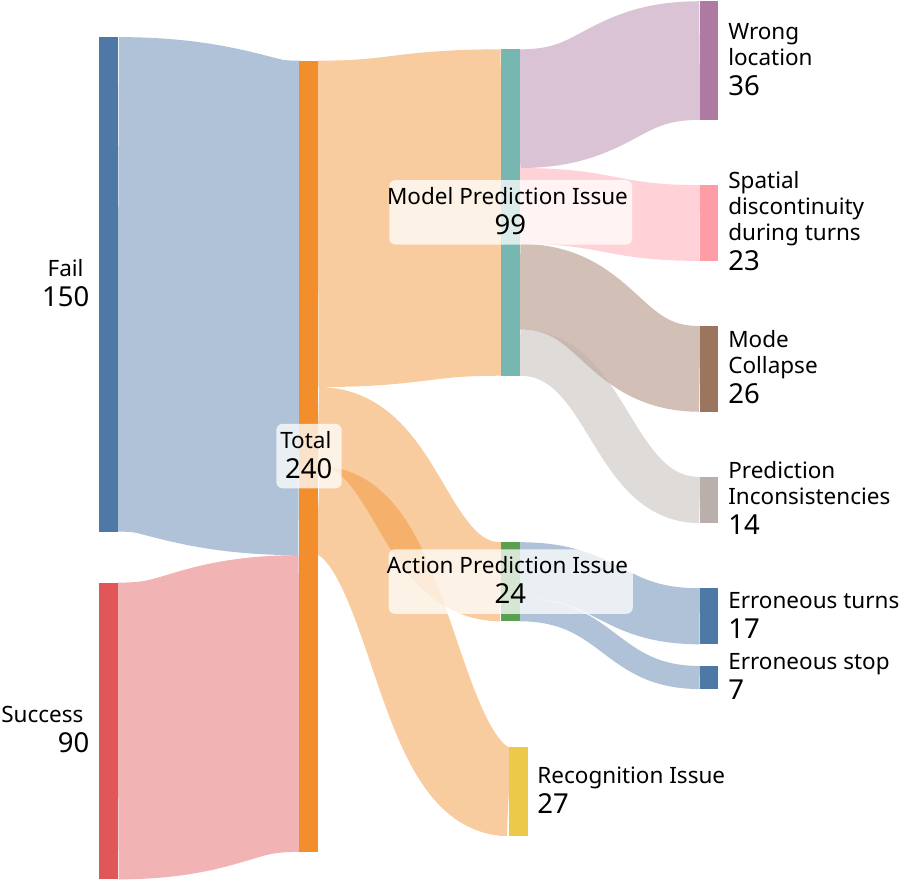}
    \caption{\textbf{Detailed analysis of the failure cases of \modelname.} }
    \label{fig:appendix_failure_case}
\end{figure*}

We conduct a comprehensive statistical analysis of the failure cases observed across all 240 real-world evaluation trajectories. As illustrated in \Cref{fig:appendix_failure_case}, out of 150 failure instances, we categorize them into three primary groups:

\begin{itemize}
    \item \textbf{Model Prediction Issues (99 cases):} This category encompasses failures stemming from the video generation process. The most prevalent error is wrong location (36 cases). This typically occurs in expansive, open environments where the target position "imagined" by the generative model misaligns with the real world (\textit{e.g.,} early turning at a certain crossing). The agent ultimately fails to localize the true destination within the context limits. The second factor is mode collapse (26 cases, occurring more frequently in night scenes), which we attribute to the currently limited training data volume (approximately 140 hours). Additionally, spatial discontinuity during turns (23 cases) occurs due to the constraints of the fixed sparse generation interval. Finally, prediction inconsistencies (14 cases) arise when the navigation intent predicted at a certain timestep misaligns with subsequent predictions during the BVN task. These observations strongly motivate our future efforts in exploraing lifelong navigation, scaling the dataset volume and introducing adaptive sparsity mechanisms to better anchor predictions in complex open worlds.
    \item \textbf{Action Prediction Issues (24 cases):} This group involves misalignments between the predicted visual future and the generated actions. Specific manifestations include erroneous turns (17 cases)—such as visually predicting a right turn but outputting a left-turn command—and erroneous stops (7 cases), where a visually predicted slight turn prematurely triggers a stop action. We hypothesize that this stems from our decoupled design choice (further discussed in Q2). While decoupling ensures critical training stability for the video generation model, relying on a separately trained lightweight action head may limit its generalization capabilities. Future work exploring an effective vision-action joint training architecture in embodied navigation could potentially alleviate this gap.
    \item \textbf{Recognition Issues (27 cases):} In some instances, VGM fails to identify the target object category, even when situated in close proximity. We attribute this representational shortcoming to the relatively small parameter size (1.3B) of our chosen base model. Future directions to resolve this include leveraging larger-scale base models coupled with advanced quantization and distillation algorithms to maintain inference efficiency.
\end{itemize}

\vspace{0.3\baselineskip}

\textbf{Q2}: \textit{What is the structural necessity of the four-stage training pipeline?}

\vspace{0.3\baselineskip}

Decoupling the video model training (Stage 2) and action training (Stage 4) is \textbf{a pragmatic design choice prioritizing training stability}.
We initially considered vision-action joint optimization. 
However, unlike robotic manipulation where the camera view remains largely static and even 1-step denoising can extract sufficient features, real-world navigation involves drastic scene variations. 
In such dynamic settings, even performing 10+ denoising steps struggles to extract usable representations, making simultaneous optimization for both high-fidelity video generation and precise action prediction highly susceptible to instability.
To ensure robust learning of the foundational video representations, we opt for this decoupled paradigm as an effective first step. 

As for adaptation (Stage 1) and distillation (Stage 3), they are served for computational efficiency (See~\Cref{sec:ablation}), which do not introduce great complexity.

\vspace{0.3\baselineskip}

\textbf{Q3}: \textit{What is the distinction between Navigation World Model~\citep{bar2025navigation} and \modelname?}

\vspace{0.3\baselineskip}

To the best of our knowledge, \modelname is the first to apply a video generation model as the primary backbone for action generation in VLN. 
The distinction between \modelname and prior Navigation World Models stems from a fundamental modality mismatch:

\begin{itemize}
    \item \textbf{Navigation World Model}: It is strictly \textbf{action-conditioned}. The architecture relies on low-level action inputs to predict the immediate next frame, rather than generating long-horizon videos.
    
    \item \textbf{\modelname}: Conversely, our framework is \textbf{language-conditioned}. It directly processes high-level language intents to achieve long-horizon video foresight, completely eliminating the need for action inputs.
\end{itemize}

Furthermore, we demonstrate that utilizing longer sequence supervision during training yields notable performance improvements (See~\Cref{sec:ablation}). This effectively validates the significance of sparse video generation in effectively extending the prediction span.

\vspace{0.3\baselineskip}

\textbf{Q4}: \textit{What about the object and language diversity of our collected real-world dataset?}

\vspace{0.3\baselineskip}

Regarding object diversity, we aim for open-world navigation. 
The targets extend far beyond ``benches or chairs'' in common simulation benchmarks, encompassing diverse \textbf{authentic physical entities} (e.g., fire hydrants, vehicles, boats, fence, elevators, road signs, bus station). 
While their locations may appear ``predictable'', they strictly follow natural real-world semantic priors. 

Regarding language diversity, we explicitly \textbf{depart from} the previous paradigm of dense instructions. 
Our motivation is to minimize the human interaction burden by using simple and high-level intents.

\vspace{0.3\baselineskip}

\textbf{Q5}: \textit{Compared to the pixel-goal guidance in InternVLA-N1~\citep{internvla-n1}, what is the advantage of sparse video guidance?}

\vspace{0.3\baselineskip}

The point-goal generated by language models are fundamentally confined to the current view.
Consequently, it often fails to assist the agent in escaping when trapped in dead ends. 
In contrast, video generation models can leverage long-horizon video guidance to imagine a backtracking trajectory to exit dead ends, thereby mitigating this perspective limitation and leading to more robust solutions for BVN tasks.

\vspace{0.3\baselineskip}

\textbf{Q6}: \textit{What is the complexity of our real-world BVN tasks?}

\vspace{0.3\baselineskip}

\begin{table}[htbp]
    \centering
    \caption{Task settings and environmental characteristics.}
    \label{tab:task_setting}
    \resizebox{\linewidth}{!}{%
    \begin{tabular}{lrll}
        \toprule
        \textbf{Task setting} & \textbf{Shortest-path distance} & \textbf{Initial target visibility} & \textbf{Environment structure} \\
        \midrule
        Indoor BVN & 8--15 m & Invisible in all tasks & Semi-open, with accessible branches \\
        \midrule
        Outdoor(Night) BVN & 15--35 m & Invisible in all tasks & Open or semi-open, with multiple accessible branches \\
        \bottomrule
    \end{tabular}%
    }
\end{table}
As shown in \Cref{tab:task_setting}, these BVN tasks are conducted in open or semi-open environments containing multiple accessible branches. The agent receives no intermediate language guidance indicating which branch to take or when to turn. If it enters an incorrect branch, it must recognize the dead-end, return to the intersection, reorient itself, and continue in the correct direction. Such exploration and recovery can add approximately 10--20 m to the executed trajectory beyond the shortest path.

Moreover, the target is invisible in the initial observation for every BVN task. In more challenging cases, the target remains unobserved for over 10-20 seconds. For reference, even when moving continuously without turning, the robot travels at most approximately 10 m within 10 seconds. The agent must therefore preserve its navigation intent and select actions over an extended period without directly observing the target.

The 256 $\text{x}$ 256 input resolution further limits the visual detail available during large turns and long-range observation. During a substantial turn, a traversable side branch may become apparent before the target or any target-relevant landmark becomes recognizable, encouraging the agent to explore an incorrect route. This difficulty is amplified at night, where low illumination and reduced contrast make distant targets and landmarks substantially harder to identify and further prolong the target-unobserved period.

Concerning the number of turns, this is policy-dependent rather than a fixed property of a BVN task because no intermediate language instruction specifies when or where the agent should turn. For example, several baselines become trapped in repeated turning near the starting position or make unexpected turns before reaching a relevant intersection.

These properties distinguish our BVN evaluation from standard IFN and many existing real-world demonstrations. To the best of our knowledge, real-world navigation tasks at this scale for the trained end-to-end model, combined with an initially invisible target, sparse high-level language, and multiple accessible branches, have rarely been evaluated systematically in existing literature. We will update the per-task distances and environment layouts in the revision.

\vspace{0.3\baselineskip}

\subsection{Task Specifications and Details}
\label{sec:task_specifications}

\begin{figure*}[h!]
    \centering
    \includegraphics[width=\linewidth]{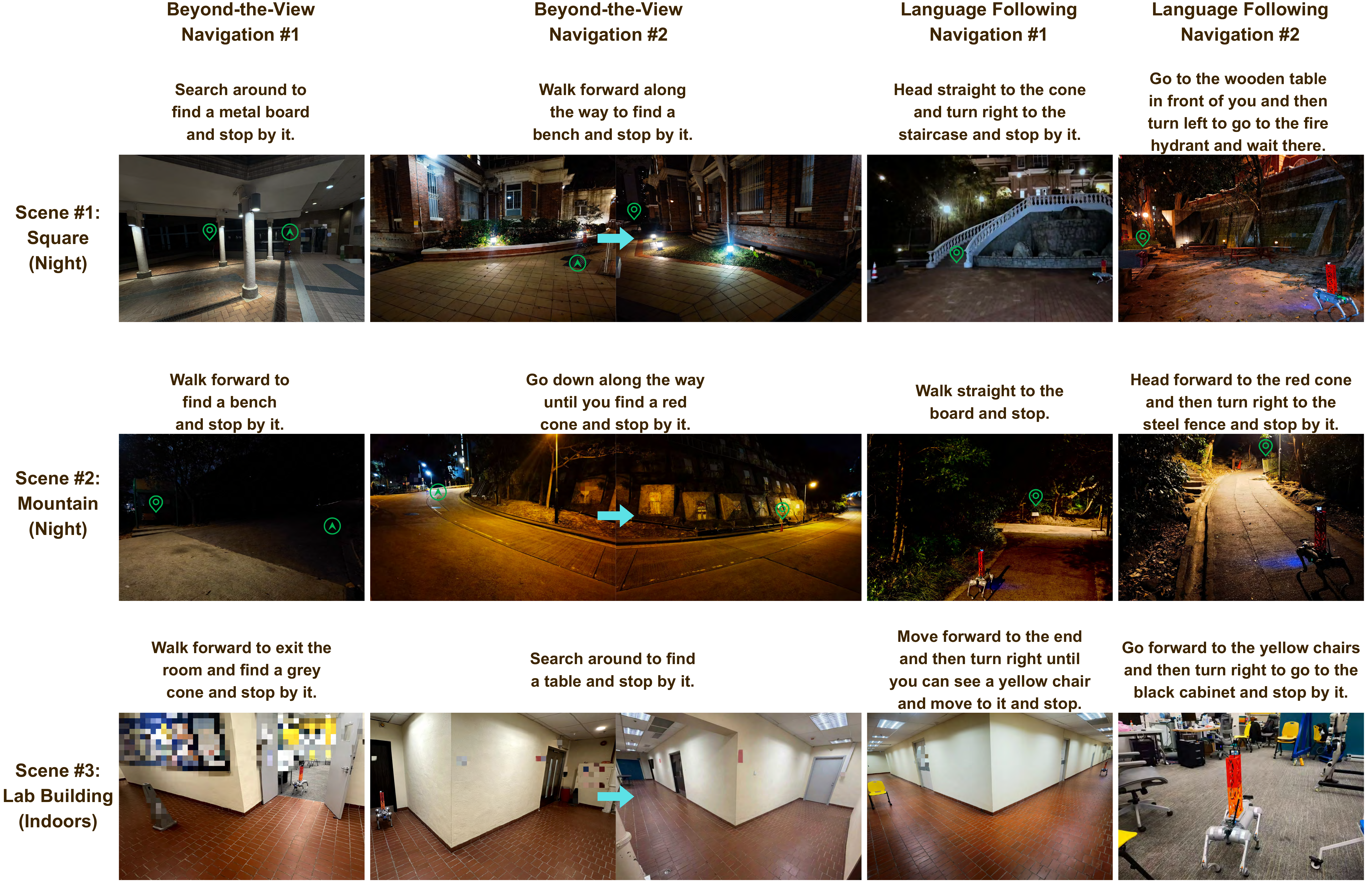}
    \caption{\textbf{Task specifications.} We present the language instructions, the initial positions of robot dog and the task settings for each scene. When the precise locations can not be clearly seen, we utilize a green arrow icon to denote the initial positions and a location pin to mark the destination for visualization clarity.}
    \label{fig:appendix_task_setting_1}
\end{figure*}

\begin{figure*}[h!]
    \centering
    \includegraphics[width=\linewidth]{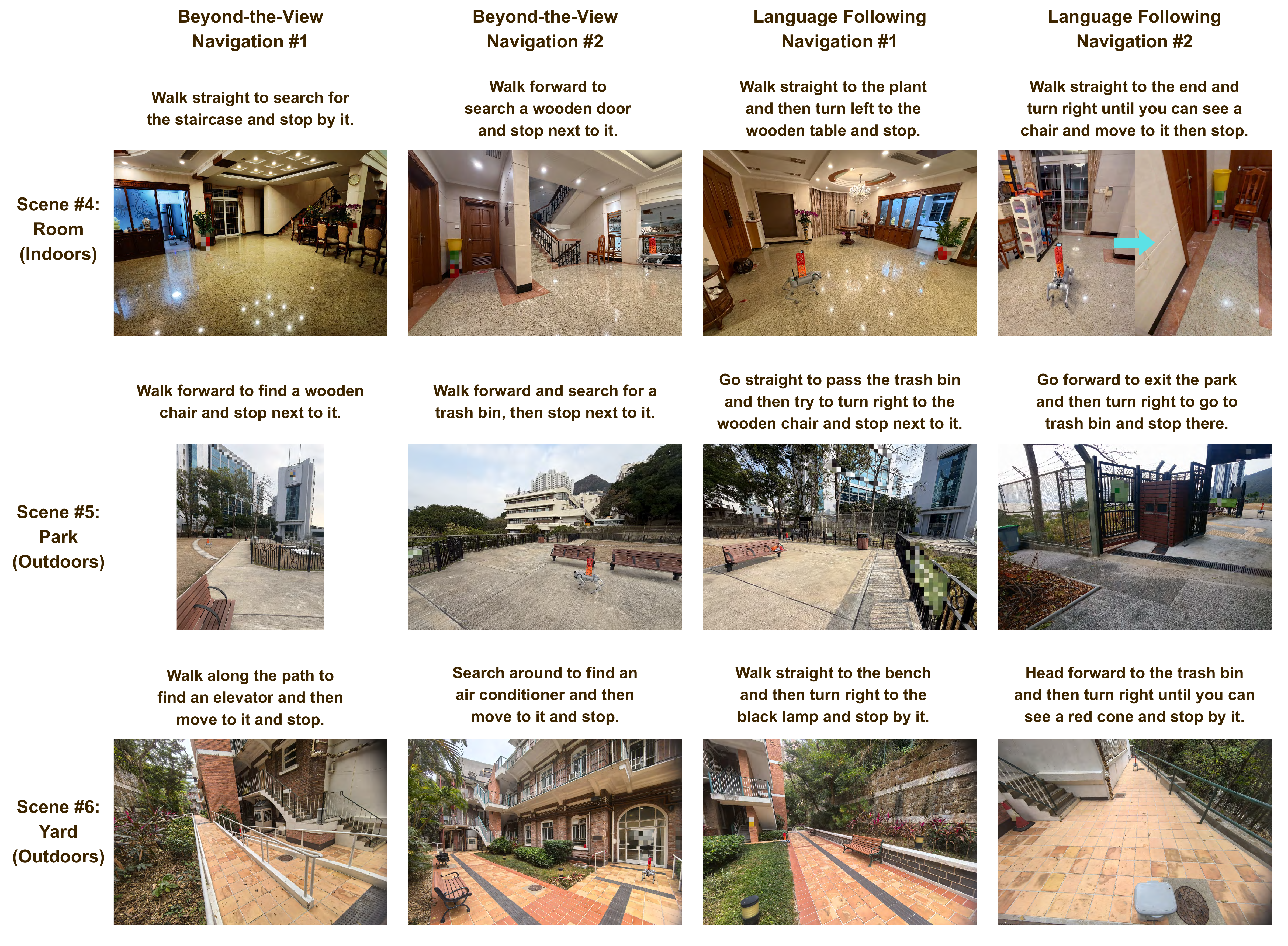}
    \caption{\textbf{Task specifications.} We present the language instructions, the initial positions of robot dog and the task settings for each scene.}
    \label{fig:appendix_task_setting_2}
\end{figure*}

\Cref{fig:appendix_task_setting_1} and~\Cref{fig:appendix_task_setting_2} provide the complete specifications for all 24 zero-shot evaluation tasks across six real-world scenes. 
Each scene contains 2 instruction-following navigation (IFN) tasks and 2 beyond-the-view navigation (BVN) tasks.

\subsection{Data Curation Details}
\label{sec:data_curation_details}

\begin{figure*}[h!]
    \centering
    \includegraphics[width=\linewidth]{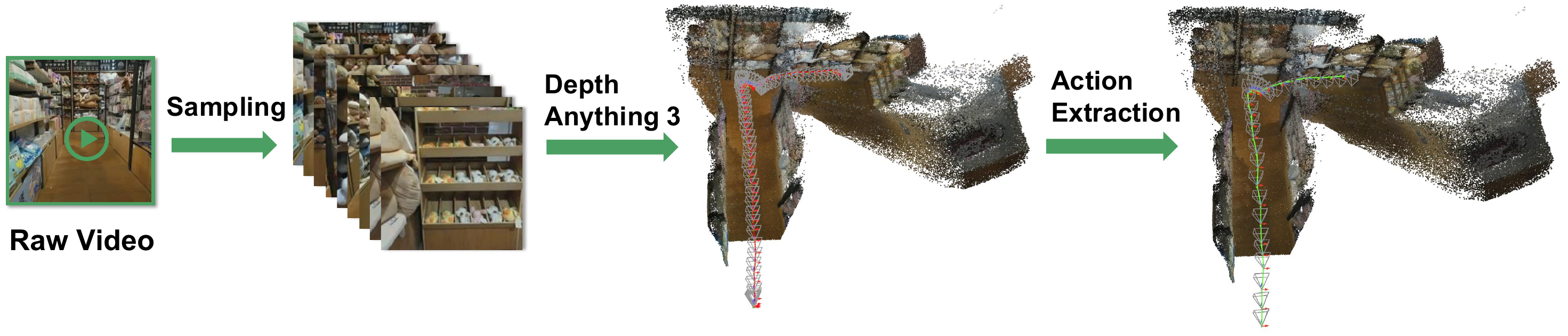}
    \caption{\textbf{Data Curation Pipeline.} The workflow consists of three stages: (1) temporal sampling from raw video; (2) pose estimation via Depth Anything 3; and (3) extrinsic-based action extraction.}
    \label{fig:appendix_data_curation}
\end{figure*}

As illustrated in Figure~\ref{fig:appendix_data_curation}, our pipeline efficiently transforms raw video streams into high-quality training image-action pairs. 
The process begins with temporal sampling, where raw videos are decimated to 4 FPS to reduce redundancy. 
We then utilize Depth Anything 3 (DA3)~\citep{depthanything3} to estimate the 6-DoF camera extrinsics for each frame, reconstructing the 3D trajectory of the agent. 
Finally, action extraction is performed by calculating the relative pose transformations between frames. 
These 3D motions are projected onto the local $XY$ plane to derive continuous action labels $(\Delta x, \Delta y, \Delta \theta)$, while static segments and views with extreme pitch angles are filtered out to ensure high-quality selection.

\subsection{Implementation Details}
\label{sec:implementation_details}

\textbf{Training Configurations.}
Our model is trained using the AdamW optimizer~\citep{loshchilov2018decoupled} on a cluster of 32 NVIDIA H200 GPUs, with the full four-stage pipeline requiring approximately 64 hours to complete. 
All training hyper-parameters of each stage is presented in Table~\ref{tab:training_hyperparams}.
% As detailed in \Cref{tab:training_hyperparams}, we carefully tune the learning rates and batch sizes for each stage, transitioning from large-scale parameter tuning in early stages to specialized fine-tuning with cosine decay in Stage 4.

\begin{table}[t!]
\centering
\caption{\textbf{Hyperparameter configurations for the four-stage training pipeline.}}
\label{tab:training_hyperparams}
\vspace{3pt}
% \resizebox{0.6\columnwidth}{!}{
%     \setlength{\tabcolsep}{1.5pt}
    \begin{tabular}{lcccccc}
    \toprule
    \textbf{Stage} & \textbf{Batch} & \textbf{Learning} & \textbf{LR} & \textbf{Trainable} & \textbf{Sampling} & \textbf{Discrete} \\
     & \textbf{Size} & \textbf{Rate} & \textbf{Scheduler} & \textbf{Params} & \textbf{Strategy} & \textbf{Steps} \\
    \midrule
    Stage 1 & 32 & $1\times10^{-5}$ & Constant & 1.3B & FM (Uniform) & 1000 \\
    Stage 2 & 32 & $1\times10^{-5}$ & Constant & 1.8B & FM (Uniform) & 1000 \\
    Stage 3 & 32 & $5\times10^{-6}$ & Constant & 1.7B & PCM (Distill) & 50 \\
    Stage 4 & 256 & $5\times10^{-5}$ & Cosine & 23.4M & DDIM & 100 \\
    \bottomrule
    \end{tabular}
% }
\end{table}

\begin{table}[t!]
\centering
\caption{\textbf{Hyper-parameters of Q-Former and Video-Former used for history compression.}}
\label{tab:qformer_arch_params}
\vspace{3pt}
% \resizebox{0.6\columnwidth}{!}{
%     \setlength{\tabcolsep}{4pt}
    \begin{tabular}{ll|ll}
    \toprule
    \multicolumn{2}{c|}{\textbf{Q-Former Configuration}} & \multicolumn{2}{c}{\textbf{Video-Former Configuration}} \\
    \midrule
    Hidden Dim & 512 & Hidden Dim & 512 \\
    Layers& 4 & Layers & 6 \\
    Heads & 8 & Heads & 8 \\
    Num. Latents & 10240 & Num. Latents & 2560 \\
    \bottomrule
    \end{tabular}
% }
\end{table}

\textbf{Network Architecture Configurations.}
Our system integrates three core components: a video generation backbone, a history compression module, and an action prediction head. 
For the video backbone (Stage 1), we directly adopt the Wan2.1 T2V-1.3B~\citep{wan2025} architecture. 
To handle high-dimensional historical context, we implement a two-stage compression strategy: a Q-Former~\citep{li2023blip} first reduces temporal redundancy, followed by a Video-Former~\citep{hu2025video} that performs $4\times$ spatial downsampling. 
This reduces the token burden to 2,560 while preserving essential context.
We present the detailed hyperparameters of Q-Former and Video-Former in Table~\ref{tab:qformer_arch_params}.

For navigation control, we design an inverse dynamics-based action head that couples a feature-aggregating Video-Former with a Diffusion Transformer (DiT)~\citep{peebles2023scalable}. 
Video-Former encodes spatiotemporal features from 640 sparse future latent tokens. 
These features then serve as the cross-attention context for the DiT to predict continuous 8-step action trajectories.
Detailed parameters for these modules are summarized in Table~\ref{tab:action_head_params}.

\begin{table}[t!]
\centering
\caption{\textbf{Hyper-parameters of the inverse dynamics-based action head}, including the Action Video-Former and the Diffusion Transformer.}
\label{tab:action_head_params}
\vspace{3pt}
% \resizebox{0.6\columnwidth}{!}{
%     \setlength{\tabcolsep}{3pt}
    \begin{tabular}{lc|lc}
    \toprule
    \multicolumn{2}{c|}{\textbf{Action Video-Former}} & \multicolumn{2}{c}{\textbf{Diffusion Transformer}} \\
    \midrule
    Hidden Dim & 256 & Hidden Dim & 256 \\
    Layers & 8 & Layers & 12 \\
    Heads & 8 & Heads & 8 \\
    Num. Latents & 640 & Num. Actions & 8 \\
    \bottomrule
    \end{tabular}
% }
\end{table}

\subsection{Hardware Deployment Details}
\label{sec:hardware_deployment_details}
We perform real-world experiments based on Unitree Go2 robotic dog.
The robot is equipped with an upward facing camera DJI Osmo Action 4 for stabilized RGB observations.
However, given that InternVLA-N1~\citep{internvla-n1} requires depth input, we adhere to its original configuration by employing an Intel® RealSense™ D455 camera, mounted with a $15^\circ$ downward pitch to acquire RGB-D observations.
The camera is secured to the back of Go2 using a custom 3D-printed bracket, positioned at a height of approximately 1m above the ground. 
This mounting height is standardized across all evaluated methods.
We deploy \modelname and all the baselines on a remote workstation with an RTX 4090 GPU.
The Go2 robot continuously sends visual observations to the server.
Upon receiving the images, the server processes them and generates corresponding navigation action commands, which are then sent back to Go2 for execution.

\begin{figure}[t]
    \centering
    \includegraphics[width=0.8\linewidth]{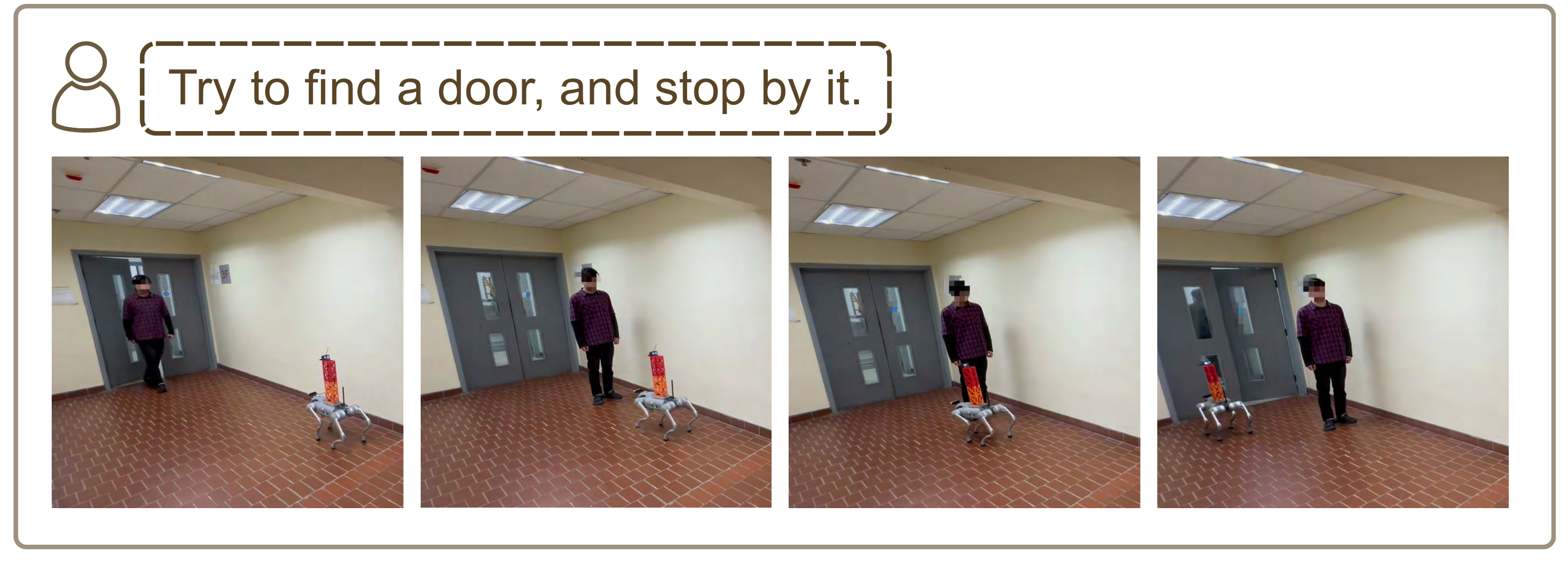}
    \caption{\textbf{Discussion
    % Qualitative results 
    of \modelname under dynamic pedestrian interference.} \modelname successfully avoids the oncoming pedestrian and reaches the door.}
    \vspace{-10pt}
    \label{fig:dynamic_avoidance}
\end{figure}

\begin{figure}[t]
    \centering
    \includegraphics[width=0.8\linewidth]{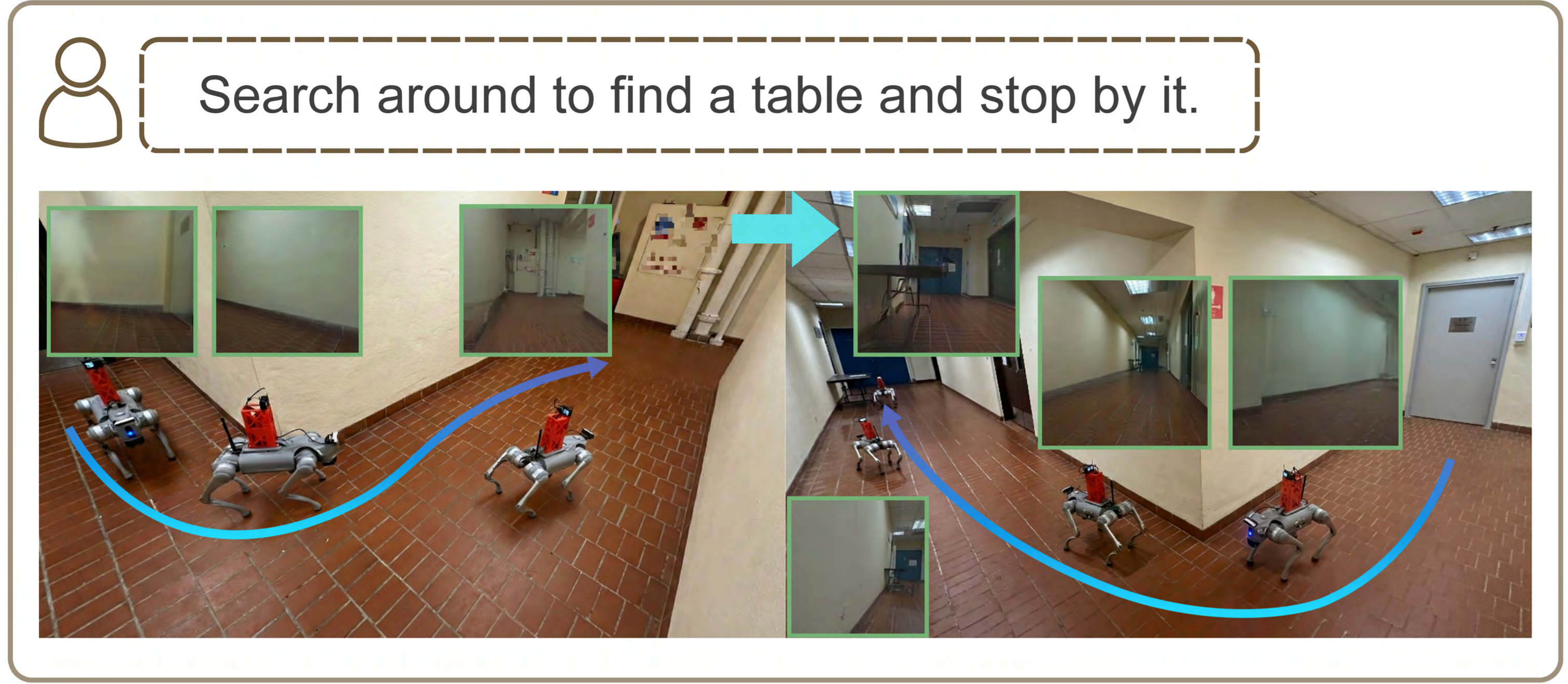}
    \caption{\textbf{Discussion
    % Qualitative results 
    of \modelname with the camera fixed at 50cm.} Images in green boxes demonstrate the predictions generated by \modelname. \modelname exhibits strong robustness against camera height variations.}
    \vspace{-10pt}
    \label{fig:low_height}
\end{figure}

\subsection{Additional Experiments}
\label{sec:further_discussion}

\noindent\textbf{Controlled Real-World-Data Comparisons.}
\label{sec:controlled_real_world}
To directly examine whether gains of \modelname primarily result from access to our 140-hour real-world dataset, we additionally fine-tuned StreamVLN, and InternVLA-N1 using the same real-world data. All entries below denote success rate (\%). “Real” denotes training or fine-tuning with our real-world navigation data.

\begin{table}[htbp]
  \centering
  \caption{Performance comparison across different scenarios against baselines finetuned by our collected real-world data.}
  \label{tab:controlled_real_world}
  \setlength{\tabcolsep}{4pt}
  \small 
  \begin{tabular}{l *{6}{c}}
    \toprule
    \textbf{Method} & \textbf{Lab IFN} & \textbf{Lab BVN} & \textbf{Yard IFN} & \textbf{Yard BVN} & \textbf{Mtn. IFN} & \textbf{Mtn. BVN} \\
    \midrule
    \textbf{SparseVideoNav} & \textbf{55} & \textbf{30} & \textbf{50} & \textbf{20} & \textbf{25} & \textbf{15} \\
    \midrule 
    StreamVLN (Real) & 40 & 10 & \textbf{50} & 0 & 20 & 0 \\
    \hline
    InternVLA-N1 (Real) & 25 & 5 & 40 & 0 & 5 & 0 \\
    \bottomrule
  \end{tabular}
\end{table}

\Cref{tab:controlled_real_world} shows the performance comparsion on baselines finetuned by our collected data. 
In BVN scenarios, we frequently observe unstable or repeated turning when the target remains outside the current view for all VLM-based baselines, consistent with the failure mode analyzed in the main paper. 
By contrast, \modelname behaves more consistently across IFN and BVN settings, including when the competing methods are given access to our real-world data. 
This controlled data-access comparison indicates that the observed improvement can be attributed to our improved model architecture.

\noindent\textbf{Dynamic Pedestrians Avoidance.}
\begin{table}[htbp]
    \centering
    \caption{Pedestrian avoidance performance across different environments.}
    \label{tab:pedestrian_avoidance}
    \resizebox{\linewidth}{!}{%
    \begin{tabular}{lccc}
        \toprule
        \textbf{Method} & \textbf{Lab pedestrian avoidance} & \textbf{Yard pedestrian avoidance} & \textbf{Mountain pedestrian avoidance} \\
        \midrule
        SparseVideoNav & 7/10 & 7/10 & 5/10 \\
        \bottomrule
    \end{tabular}
    }
\end{table}
\begin{table}[htbp]
    \centering
    \caption{Performance evaluation under different camera heights.}
    \label{tab:camera_height}
    \resizebox{\linewidth}{!}{%
    \begin{tabular}{lcccccc}
        \toprule
        \textbf{Camera height} & \textbf{Lab IFN} & \textbf{Lab BVN} & \textbf{Yard IFN} & \textbf{Yard BVN} & \textbf{Mountain IFN} & \textbf{Mountain BVN} \\
        \midrule
        SparseVideoNav (1.0 m) & 55 & 30 & 50 & 20 & 25 & 15 \\
        SparseVideoNav (0.7 m) & 50 & 20 & 55 & 20 & 15 & 5 \\
        \bottomrule
    \end{tabular}%
    }
\end{table}
\begin{table}[!ht]
    \centering
    \caption{Success Rate on longer cross-room and multi-turn evaluation tasks, with \textit{prompt: Walk exit the room to the yellow chair and then turn right to go to the black box, and then turn left to enter the lab to get to the metal cart and stop.}} 
    \label{tab:cross_room_results}
    \begin{tabular}{lc}
        \toprule
        \textbf{Method} & \textbf{SR} \\
        \midrule
        \textbf{SparseVideoNav} & \textbf{4/10} \\
        \midrule
        CorrectNav(Real) & 1/10 \\
        OmniNav(Real) & 3/10 \\
        StreamVLN(Real) & 2/10 \\
        \bottomrule
    \end{tabular}
\end{table}
\begin{table}[!ht]
    \centering
    \caption{Quantitative ablation of the sparse interval.}
    \label{tab:sparse_interval}
    \resizebox{\linewidth}{!}{%
    \begin{tabular}{lcccccc}
        \toprule
        \textbf{Sparse interval} & \textbf{Lab IFN} & \textbf{Lab BVN} & \textbf{Yard IFN} & \textbf{Yard BVN} & \textbf{Mountain IFN} & \textbf{Mountain BVN} \\
        \midrule
        1 & \textbf{65} & 10 & \textbf{50} & 10 & \textbf{30} & 0 \\
        \textbf{3} & 55 & \textbf{30} & \textbf{50} & \textbf{20} & 25 & \textbf{15} \\
        5 & 30 & 5 & 25 & 5 & 5 & 0 \\
        \bottomrule
    \end{tabular}%
    }
\end{table}
\begin{table}[!ht]
    \centering
    \caption{Ablation on distillation steps.}
    \label{tab:distillation_steps}
    \resizebox{\linewidth}{!}{%
    \begin{tabular}{lccccccr}
        \toprule
        \textbf{Distillation steps} & \textbf{Lab IFN} & \textbf{Lab BVN} & \textbf{Yard IFN} & \textbf{Yard BVN} & \textbf{Mountain IFN} & \textbf{Mountain BVN} & \textbf{Inference latency} \\
        \midrule
        M = 2 & 10 & 0 & 5 & 5 & 0 & 0 & \textbf{0.48 s} \\
        \textbf{M = 4} & 55 & 30 & \textbf{50} & \textbf{20} & \textbf{25} & 15 & 0.79 s \\
        M = 8 & \textbf{60} & \textbf{35} & \textbf{50} & \textbf{20} & 20 & \textbf{20} & 1.41 s \\
        \bottomrule
    \end{tabular}%
    }
\end{table}
\begin{table}[!ht]
  \centering
  \caption{Performance comparison with different action heads across scenarios.}
  \label{tab:action_head_comparison}
  \resizebox{\linewidth}{!}{
    \begin{tabular}{@{}lcccccc@{}}
      \toprule
      \textbf{Action head} & \textbf{Lab IFN} & \textbf{Lab BVN} & \textbf{Yard IFN} & \textbf{Yard BVN} & \textbf{Mountain IFN} & \textbf{Mountain BVN} \\
      \midrule
      \textbf{Diffusion} & \textbf{55} & \textbf{30} & \textbf{50} & \textbf{20} & \textbf{25} & \textbf{15} \\
      MLP regression & 20 & 5 & 35 & 10 & 5 & 5 \\
      \bottomrule
    \end{tabular}%
  }
\end{table}

Given that DA3~\citep{depthanything3} struggles with reliable action estimation in the presence of frontal dynamic pedestrians, we filter out such trajectories to preserve the accuracy of the actions.
Remarkably, despite this exclusion, \modelname exhibits an emergent capability to dynamically avoid pedestrians during deployment, underscoring its strong adaptability and generalization, as shown in \Cref{fig:dynamic_avoidance} and \Cref{tab:pedestrian_avoidance}.

\noindent\textbf{Camera Height Insensitivity.}
We further observe that while LLM-based paradigm is notably susceptible to camera height shifts, video generation-based paradigm maintains robust against these discrepancies.
Despite being trained on data collected at an approximate height of 1m, \modelname successfully performs navigation with a fixed camera height of 50cm, as shown in \Cref{fig:low_height} and \Cref{tab:camera_height}.

\noindent\textbf{Longer Cross-Room and Multi-Turn Evaluation Tests.}
We further conducted an additional evaluation on three-stage IFN setting.
As shown in~\ref{tab:cross_room_results}, SparseVideoNav succeeds in 4/10 trials, compared with 1/10 for CorrectNav, 3/10 for OmniNav, and 2/10 for StreamVLN. It shows that SparseVideoNav remains effective on longer trajectories requiring multiple sequential turns. 

\noindent\textbf{Sparse Interval.} 
We further conducted an additional evaluation on sparse interval. 
As shown in \Cref{tab:sparse_interval}, when interval equals to 1, it is beneficial for IFN, but the short supervision coverage reduces the performance of BVN. 
As the interval gradually increases, artifacts easily arise during the continuous movement, leading to a performance drop for both IFN and BVN. 
Thus, interval equals to 3 is a favorable choice.

\noindent\textbf{Denoising Steps.}
As shown in \Cref{tab:distillation_steps}, reducing the number of steps to $M = 2$ improves inference speed but degrades the generated future representations and downstream navigation performance. 
Increasing it to $M = 8$ improves generation quality only marginally while increasing inference latency. 
The results therefore indicate that $M = 4$ provides a suitable balance between navigation performance and real-time inference efficiency.

\noindent\textbf{Action Head.}
Our motivation for using a diffusion-based action head follows prior findings in manipulation: diffusion policies can feature multimodal action distributions, whereas deterministic regression tends to average multiple feasible action modes.

This property is relevant to navigation because the same high-level intent and predicted future can admit multiple valid local trajectories, particularly near open intersections or during obstacle avoidance. We added a direct comparison against an MLP regression head while keeping the VGM backbone and training data unchanged.

As shown in~\Cref{tab:action_head_comparison}, the diffusion-based action head consistently performs better than direct MLP regression.

\subsection{Data Ethics and License}
\label{sec:data_ethics}

% In this section, we detail the ethical considerations and privacy-preserving measures taken during the curation of our 140-hour real-world VLN dataset.

\textbf{Privacy Protection and De-identification.} 
To strictly protect individual privacy, we implement a robust and irreversible de-identification processing pipeline. 
All captured video frames are processed by human expert to identify all sensitive information. 
All sensitive information, including human faces and vehicle identifiers, is permanently blurred before being used for model training or stored in our servers. 
We ensure that no identifiable personal information is present in the final dataset.

\textbf{Data Collection Ethics.} 
Our data collection involves human operators recording in diverse public environments. 
All operators are briefed on ethical guidelines, ensuring that they follow local regulations and avoid recording in restricted or private areas without explicit authorization. 
Data collection is performed using a handheld DJI Osmo Action 4 in real-world environments. This manual approach ensures data diversity and procedural transparency throughout the recording process.

\textbf{License.}
We build upon Wan2.1~\citep{wan2025}, which is under Apache License 2.0.
The code and dataset for \modelname would be open-sourced under CC BY-NC-SA 4.0 License.

%% file: checklist.tex
\section*{NeurIPS Paper Checklist}

\begin{enumerate}

\item {\bf Claims}
    \item[] Question: Do the main claims made in the abstract and introduction accurately reflect the paper's contributions and scope?
    \item[] Answer: \answerYes{} % Replace by \answerYes{}, \answerNo{}, or \answerNA{}.
    \item[] Justification: Yes, the main claims are clearly laid out at the end of introduction and validated in~\Cref{sec:exp}.

\item {\bf Limitations}
    \item[] Question: Does the paper discuss the limitations of the work performed by the authors?
    \item[] Answer: \answerYes{} % Replace by \answerYes{}, \answerNo{}, or \answerNA{}.
    \item[] Justification: Yes, see the end of~\cref{sec:conclusion} and our analysis of failure cases in~\Cref{sec:motivating_questions}.

\item {\bf Theory assumptions and proofs}
    \item[] Question: For each theoretical result, does the paper provide the full set of assumptions and a complete (and correct) proof?
    \item[] Answer: \answerNA{} % Replace by \answerYes{}, \answerNo{}, or \answerNA{}.
    \item[] Justification: This paper does not have any theoretical results.

    \item {\bf Experimental result reproducibility}
    \item[] Question: Does the paper fully disclose all the information needed to reproduce the main experimental results of the paper to the extent that it affects the main claims and/or conclusions of the paper (regardless of whether the code and data are provided or not)?
    \item[] Answer: \answerYes{} % Replace by \answerYes{}, \answerNo{}, or \answerNA{}.
    \item[] Justification: Yes, see our appendix for training details, implementation details, and hardware deployment details.

\item {\bf Open access to data and code}
    \item[] Question: Does the paper provide open access to the data and code, with sufficient instructions to faithfully reproduce the main experimental results, as described in supplemental material?
    \item[] Answer: \answerYes{} % Replace by \answerYes{}, \answerNo{}, or \answerNA{}.
    \item[] Justification: Please see the link below the title.

\item {\bf Experimental setting/details}
    \item[] Question: Does the paper specify all the training and test details (e.g., data splits, hyperparameters, how they were chosen, type of optimizer) necessary to understand the results?
    \item[] Answer: \answerYes{} % Replace by \answerYes{}, \answerNo{}, or \answerNA{}.
    \item[] Justification: Yes, see our appendix for training details, implementation details, and hardware deployment details.

\item {\bf Experiment statistical significance}
    \item[] Question: Does the paper report error bars suitably and correctly defined or other appropriate information about the statistical significance of the experiments?
    \item[] Answer: \answerNo{} % Replace by \answerYes{}, \answerNo{}, or \answerNA{}.
    \item[] Justification: A common practice in the field of embodied navigation is to evaluate models and directly report the performance.

\item {\bf Experiments compute resources}
    \item[] Question: For each experiment, does the paper provide sufficient information on the computer resources (type of compute workers, memory, time of execution) needed to reproduce the experiments?
    \item[] Answer: \answerYes{} % Replace by \answerYes{}, \answerNo{}, or \answerNA{}.
    \item[] Justification: Yes, see our experiments for the details.
    
\item {\bf Code of ethics}
    \item[] Question: Does the research conducted in the paper conform, in every respect, with the NeurIPS Code of Ethics \url{https://neurips.cc/public/EthicsGuidelines}?
    \item[] Answer: \answerYes{} % Replace by \answerYes{}, \answerNo{}, or \answerNA{}.
    \item[] Justification: We make sure to preserve anonymity and conform to the NeurIPS Code of Ethics.

\item {\bf Broader impacts}
    \item[] Question: Does the paper discuss both potential positive societal impacts and negative societal impacts of the work performed?
    \item[] Answer: \answerYes{} % Replace by \answerYes{}, \answerNo{}, or \answerNA{}.
    \item[] Justification: The positive impacts are discussed in~\Cref{sec:intro}.
    
\item {\bf Safeguards}
    \item[] Question: Does the paper describe safeguards that have been put in place for responsible release of data or models that have a high risk for misuse (e.g., pre-trained language models, image generators, or scraped datasets)?
    \item[] Answer: \answerYes{} % Replace by \answerYes{}, \answerNo{}, or \answerNA{}.
    \item[] Justification: Yes, we will open-source our dataset after processing it with de-identification pipeline.

\item {\bf Licenses for existing assets}
    \item[] Question: Are the creators or original owners of assets (e.g., code, data, models), used in the paper, properly credited and are the license and terms of use explicitly mentioned and properly respected?
    \item[] Answer: \answerYes{} % Replace by \answerYes{}, \answerNo{}, or \answerNA{}.
    \item[] Justification: Yes, see~\Cref{sec:data_ethics}. 

\item {\bf New assets}
    \item[] Question: Are new assets introduced in the paper well documented and is the documentation provided alongside the assets?
    \item[] Answer: \answerYes{} % Replace by \answerYes{}, \answerNo{}, or \answerNA{}.
    \item[] Justification: Yes, we will open-source our dataset alongside a thorough usage document.

\item {\bf Crowdsourcing and research with human subjects}
    \item[] Question: For crowdsourcing experiments and research with human subjects, does the paper include the full text of instructions given to participants and screenshots, if applicable, as well as details about compensation (if any)? 
    \item[] Answer: \answerNA{} % Replace by \answerYes{}, \answerNo{}, or \answerNA{}.
    \item[] Justification: The paper does not involve crowdsourcing nor research with human subjects.

\item {\bf Institutional review board (IRB) approvals or equivalent for research with human subjects}
    \item[] Question: Does the paper describe potential risks incurred by study participants, whether such risks were disclosed to the subjects, and whether Institutional Review Board (IRB) approvals (or an equivalent approval/review based on the requirements of your country or institution) were obtained?
    \item[] Answer: \answerNA{} % Replace by \answerYes{}, \answerNo{}, or \answerNA{}.
    \item[] Justification: The paper does not involve crowdsourcing nor research with human subject.

\item {\bf Declaration of LLM usage}
    \item[] Question: Does the paper describe the usage of LLMs if it is an important, original, or non-standard component of the core methods in this research? Note that if the LLM is used only for writing, editing, or formatting purposes and does \emph{not} impact the core methodology, scientific rigor, or originality of the research, declaration is not required.
    %this research? 
    \item[] Answer: \answerNA{} % Replace by \answerYes{}, \answerNo{}, or \answerNA{}.
    \item[] Justification: LLM is used only for polishing purpose.

\end{enumerate}